\documentclass[3p,a4paper]{elsarticle}

\usepackage{amssymb,amsmath,amstext,amscd}

\usepackage{booktabs}
\usepackage{multirow}
\usepackage{subfig}
\usepackage{graphicx}
\usepackage{xcolor}
\usepackage{hyperref}
\usepackage{hypcap}
\usepackage{underoverlap}
\usepackage[utf8]{inputenc}
\usepackage[T1]{fontenc}
\usepackage{natbib}
\usepackage{longtable}
\journal{Applied Soft Computing, Elsevier}

\begin{document}

\begin{frontmatter}

%\title{Elsevier \LaTeX\ template\tnoteref{mytitlenote}}

\title{Scalable and Customizable Benchmark Problems for Many-Objective Optimization}
% \tnotetext[mytitlenote]{Fully documented templates are available in the elsarticle package on \href{http://www.ctan.org/tex-archive/macros/latex/contrib/elsarticle}{CTAN}.}

\author[ifmg]{Ivan Reinaldo Meneghini}
\ead{ivan.reinaldo@ifmg.edu.br}

\author[ppgee,minds]{Marcos Antonio Alves}
\ead{marcosalves@ufmg.br}

\author[uminho]{António Gaspar-Cunha}
\ead{agc@dep.uminho.pt}

\author[minds]{Frederico Gadelha Guimarães\corref{mycorrespondingauthor}}
\cortext[mycorrespondingauthor]{Corresponding author}
\ead{fredericoguimaraes@ufmg.br}
\ead[url]{https://minds.eng.ufmg.br/}

\address[ifmg]{Instituto Federal de Educação, Ciência e Tecnologia de Minas Gerais (IFMG) -- Campus Ibirité, Minas Gerais, Brazil}

\address[ppgee]{Graduate Program in Electrical Engineering - Federal University of Minas Gerais, Av. Antônio Carlos 6627, 31270-901, Belo Horizonte, MG, Brazil}

\address[minds]{Machine Intelligence and Data Science (MINDS) Laboratory, Federal University of Minas Gerais (UFMG), Av. Antônio Carlos 6627, 31270-901, Belo Horizonte, MG, Brazil}

\address[uminho]{Institute of Polyners and Composites, University of Minho (Uminho), campus Azurém, 4800-058 Guimarães, Portugal}


\begin{abstract}
Solving many-objective problems (MaOPs) is still a significant challenge in the multi-objective optimization (MOO) field. One way to measure algorithm performance is through the use of  benchmark functions (also called test functions or test suites), which are artificial problems with a well-defined mathematical formulation, known solutions and a variety of features and difficulties.
In this paper we propose a parameterized generator of scalable and customizable benchmark problems for MaOPs. It is able to generate problems that reproduce features present in other benchmarks and also problems with some new features. We propose here the concept of generative benchmarking, in which one can generate an infinite number of MOO problems, by varying parameters that control specific features that the problem should have: scalability in the number of variables and objectives, bias, deceptiveness, multimodality, robust and non-robust solutions, shape of the Pareto front, and constraints. The proposed Generalized Position-Distance (GPD) tunable benchmark generator uses the position-distance paradigm, a basic approach to building test functions, used in other benchmarks such as Deb, Thiele, Laumanns and Zitzler (DTLZ), Walking Fish Group (WFG) and others. It includes scalable problems in any number of variables and objectives and it presents Pareto fronts with different characteristics.
The resulting functions are easy to understand and visualize, easy to implement, fast to compute and their Pareto optimal solutions are known.
\end{abstract}


\end{frontmatter}

%\linenumbers

%%=========================================
%%=========================================
\section{Introduction}
\label{sec_introduction}

One significant challenge in the multi-objective optimization (MOO) field is related to solving many-objective problems (MaOPs), which are usually defined when the number of objective functions is greater than three \cite{jain2014evolutionary,cheng2017benchmark,zhou2017ensemble,maltese2018scalability}. The increase in the number of objectives poses a number of challenges to the methods designed for MOO, in terms of convergence to the Pareto optimal solutions, dimensionality of the Pareto front, visualization of solutions and decision-making. An efficient way of obtaining an approximation set of the solutions to these problems is through stochastic heuristic algorithms, particularly Multi-Objective Evolutionary Algorithms (MOEA) \cite{li2015many,ishibuchi2008reviewmaop}. Assuming no prior preference provided by the decision-maker (DM), MOEAs are designed to find an unbiased, well-distributed approximation of the entire Pareto front, a task that becomes harder in MaOPs \cite{meneghini-eurogen-2019}. It is also possible to use preference information in MOEAs \cite{meneghini-eurogen-2019,cheng2016reference,goulart2016preference,thiele2009preference}. 
Nonetheless, although focusing the search on a given region of interest, the scalability issue remains a challenge for preference-based MOEA.

Several factors characterize a good approximation set of solutions to a MOO problem:  
%A solution must, first of all, be compatible with the characteristics and limitations of the proposed problem to be implemented. The feasibility is easily verified: the solutions obtained need to satisfy the problem constraints, both in the decision space and in the objective space. 
convergence to the true Pareto front, representativeness of the set (also involving the concept of diversity) and coverage of the obtained approximation in the objective space. Coverage is generally understood as the extension of the set or how well the set of solutions covers the extreme points of the Pareto front. Representativeness is the presence and good distribution of solutions along the Pareto front surface, providing to the DM a good representation of the Pareto front in terms of the potential to analyzing different trade-offs related to the objectives. Convergence means that the solutions obtained should be as close as possible to the Pareto front. %This factor is usually measured by metrics such as the Generational Distance (GD) \cite{VanVeldhuizen1999}, Inverted Generational Distance (IGD) \cite{Czyzzak1998}) and Hypervolume indicator (HV) \cite{Zitzler:1998}. However, either they are extremely dependent on the sample of points in the true Pareto front (in case of GD and IGD) or have a very high computational cost (HV) when the number of objectives increase.
Another factor that may be desirable in many practical cases is related to the robustness of the solutions found. In real-world problems the presence of noise, disturbances and variability is a rule, not an exception. In addition to being locally Pareto optimal, it is desirable that the solutions offered to the DM are less sensitive to the impact of uncertainties or unforeseen scenarios. In this way a sub-optimal solution (locally Pareto optimal), but that presents little variation under the presence of noise and uncertainties, can be considered better than an optimal solution that presents a high variability due to these effects.

%An efficient MOEA may find feasible solutions with good convergence and dispersion, using adequate computational and time resources. A robust MOEA seeks, in addition to feasibility, convergence and dispersion, solutions that have the lowest sensitivity to noise factors and uncertainties. 
One way to measure algorithm performance is through the use of benchmark functions (also called test functions or test suites in the literature), which are artificial problems with a well-defined mathematical formulation, known solutions and a variety of features and difficulties. Benchmarking allows one to test algorithms in the task of obtaining approximations to Pareto fronts of these test functions, such that the quality of these approximations, and hence the performance of the algorithm, can be measured. The key assumption behind this is that an optimization algorithm that performs well in those problems would also perform well in real-world problems. Additionally, by analyzing the performance of a given algorithm or group of algorithms in problems with different features, the designer can better understand the strenghts and weaknesses of each method. The use of benchmarking drives research  and development in computational intelligence, optimization and machine learning, allowing to find the weaknesses and strengths of MOEAs more comprehensively \cite{wang2018generator}. %the evaluation and comparison of different algorithms, expanding applicability and pushing the limits of performance in the algorithms developed and used by the community.   

%The specialized literature presents a great variety of problems \citep{Hui_Li_2009,Ishibuchi2017,GasparCunha.etal:2013,cheng2017benchmark}. This work will focus their attention on the analysis in three sets of test problems: ZDT, DTLZ and WFG. The choice of test functions is justified by their wide use in specialized literature. 

There are many benchmark problems available for assessing the performance of MOEA, such as those described by \citet{cheng2017benchmark} and \citet{tian2017platemo}. Some of the well known and widely used test problems are DTLZ \cite{Deb2005DTLZ}, WFG \cite{huband2006review,huband2005scalable}, ZDT \cite{zitzler2000ZDT}, CTP \cite{deb2001CTP} among others, see \cite{tian2017platemo}. 
%This work will focus on the analysis of three popular sets of test problems: ZDT, DTLZ and WFG. The choice of test functions is justified by their wide use in specialized literature. 
Some of these benchmark problems have been repeatedly used for demonstrating difficulties of the Pareto dominance based MOEA algorithms in MaOPs, see for instance \cite{ishibuchi2017performance}. 
%However, these problems present some weak points such as very simple Pareto fronts, either they are flat, spheres, curves or a simplex, these problems do not consider robustness and inequality/equality constraints, few of them are able to deal satisfactorily with many objectives or have a known Pareto set of solutions or yet, are interpretable. 
More recently, many studies have identified weaknesses with the most well-known benchmark problems, pointing out that new test problems are desirable in the literature to drive research and development of MOEA \cite{cheng2017benchmark,wang2018generator}. Some studies have tried to identify desirable characteristics of benchmark problems in MaOPs, see for instance \cite{zapotecas2019review}, many of which are not present in most benchmarks in the literature. More recent benchmark problems have been proposed in \cite{wang2018generator,matsumoto2019multiobjective,yue2019novel,jiang2019scalable,yujinolhofer2019benchmark,mawang2019evolutionary,weise2018difficult}. Benchmark problems are discussed in Section \ref{sec_reviewtestproblems} and the desirable characteristics in Section \ref{sub_desirablecharacteristics}.
%On the one hand, the ineffectiveness of Pareto dominance relation, most important criterion in MOO, has led to an underwhelming performance of traditional Pareto-based algorithms \cite{cheng2017benchmark}. On the other hand, some algorithms are overspecialized in some test problems \cite{ishibuchi2017performance}. 

Nowadays, with the variety of benchmarks available in the literature, researchers got to a point in which any new MOEA developed should be tested on a wide variety of benchmark test suites, with dozens of different problems.  Instead of proposing yet another set of fixed benchmark problems, this paper takes a different stand. In this paper we propose a parameterized generator of scalable and customizable benchmark problems for MaOPs. It is able to generate problems that reproduce features present in other benchmarks and also problems with some new features. The software engineering community has developed some years ago the concept of generative testing. In short, generative testing allows one to specify properties the software should have. Then the testing library generates test cases in a smart way \cite{dias2007survey,Pires2018}. Following this idea, we propose here the concept of generative benchmarking, which is a similar approach for benchmarking MOEAs: we develop in this paper a test generator, able to generate an infinite number of MOO problems, by varying a number of parameters that control specific features that the problem should have.

With this generative testing approach, one can generate scalable and customizable benchmark problems by controlling a number of features, such as scalability in the number of variables and objectives, bias, robustness, deceptiveness, multimodality, shape of the Pareto front, and constraints. These features are discussed in more detail in Section \ref{sec_newfamiltybenchproblems}. The instance generator (see Section \ref{sec_proposedfunctions}) offers the possibility of precise control over the spatial location of points in the objectives space, which is an essential characteristic to verify the efficiency of methods that propose to find solutions in specific regions of the space of the objectives, mainly in MaOPs. This work opens up a new perspective in benchmarking and testing MOEA. A number of test cases, combining different characteristics, can be randomly and automatically generated and the competing algorithms are then executed over each test case. Nonparametric statistical analysis for multiple comparison can then be used, following the best practices in experimental comparison of stochastic algorithms \cite{garcia2010advanced,derrac2011practical,trawinski2012nonparametric}.

The paper is organized as follows: Section \ref{sec_reviewtestproblems} presents an overview of benchmark problems in the literature as well as  their limitations. Section \ref{sec_newfamiltybenchproblems} describes desirable characteristics for test problems and introduces a new set of benchmark functions that pose more challenges for MOEA. 
Section \ref{sec_proposedfunctions} describes the parameterized generator of scalable and customizable benchmark problems.
At the end, Section \ref{sec_conclusions} describes the conclusions and discusses some possible directions for future research. 

\section{An Overview of Test Functions}
\label{sec_reviewtestproblems}

In this section, we review benchmark problems in the literature for multi-objective test problems: Zitzler, Deb and Thiele (ZDT) test suite \cite{zitzler2000ZDT}, Deb, Thiele, Laumanns and Zitzler (DTLZ)  test suite \cite{Deb2005DTLZ}, and Walking Fish Group (WFG) toolkit \cite{huband2006review}. A common feature of these problems is that the optimization variables can always be written as $\mathbf{x}=(\mathbf{x}_p,\mathbf{x}_d)$, where $\mathbf{x}_p$ is a vector with $M-1$ elements that controls the position of its image in the objective space and $\mathbf{x}_d$ is a vector with  $N - (M-1)$ coordinates that controls the distance of its image to the Pareto front.
Recent benchmarks in the literature are discussed in Section \ref{subsec_recent}. Finally, in Section \ref{subsec_limitationstestfunctions} we discuss their main limitations.

%%=========================================
\subsection{ZDT benchmark}
\label{subsec_zdt}

Zitzler, Deb and Thiele (ZDT) \cite{zitzler2000ZDT} proposed the family of functions ZDT1-ZDT6, in order to compare the effectiveness of different MOEAs. Although this benchmark considers only problems with two objectives, it introduced the idea of having a subset of variables responsible for the position and variables responsible for the distance in the objective space. This constructive approach for benchmark problems is used in other works that followed on benchmarking for MaOPs.

The ZDT problems have variables in the $[0,1]$ range, except for ZDT5, which has binary domain \cite{huband2006review}. In all functions, $\mathbf{x}_p$ is composed by only one coordinate, which corresponds to the objective value $f_1(\mathbf{x})=x_1$. The second objective $f_2(\mathbf{x})$ is defined by means of different expressions with $\mathbf{x}_p$ and $\mathbf{x}_d$, being responsible for convergence and distribution of the population on the Pareto front in the different problems. Vector $\mathbf{x}_d$ has a variable size (29 for ZDT1 to ZDT3, 9 for ZDT4 and ZDT6 and 10 for ZDT5). The functions present Pareto fronts that are convex (ZDT1 and ZDT4), concave (ZDT2 and ZDT6) and disconnect (ZDT3) \cite{huband2006review}. 
However, the ZDT benchmark is limited to two objectives and its main focus is on the convergence of the solutions towards the Pareto front. %The authors still highlight Pareto front coverage and the maintenance of population diversity which characterize the challenges to be presented for MOEAs. 

%%=========================================
\subsection{DTLZ benchmark}
\label{subsec_dtlz}

%The DTLZ family of test functions appears in two identical technical reports in 2001 (Scalable Test Problems for Evolutionary Multi-Objective Optimization) published in the Technical Report No. 112 of the Swiss Federal Institute of Technology (ETH) and in the KanGAL Report Number 2001001 of the Indian Institute of Technology Kanpur, in a paper of the IEEE Congress on Evolutionary Computation (CEC) in 2002 (Scalable multi-objective optimization test problems) and as an article of the magazine Evolutionary Multiobjective Optimization: Theoretical Advances and Applications in the year 2005 under the title Scalable Test Problems for Evolutionary Multiobjective Optimization. The paper presented in IEEE CEC 2002 is a summarized version of the technical reports, with the withdrawal of the DTLZ5, DTLZ8 and DTLZ9 problems and the reordering of the remaining problems \footnote{The problems DTLZ5 to DTLZ7 presented at the congress correspond to problems DTLZ6 to DTLZ8 of the technical reports.} The difference between the technical reports and the 2005 article is the presentation in the 2005 article of the comet problem in the appendix. This problem was present as a section in the development of the ideas presented in the technical reports (section 4.6), which was fully transposed to the appendix. In this paper, the authors highlight the following desirable characteristics for a test problem:

Deb, Thiele, Laumanns, and Zitzler (DTLZ) \cite{Deb2005DTLZ} presented the DTLZ1-DTLZ9 problems, which are scalable for any number of objectives. Scalability is a desirable feature which makes these test functions suitable for testing MOEA in MaOPs \cite{Deb2005DTLZ,huband2006review}. These problems have a well-defined solution, namely $ x_i \in [0,1] $ for $ x_i \in \mathbf{x}_p $ and $ x_j = 0.5 $ for $ x_j \in \mathbf{x}_d $. In all problems, the Pareto front is located in the first orthant\footnote{The first orthant is the set of points $ \mathbf{x} = (x_1, \ldots, x_M) $ of the $ M-$dimensional space with $ x_j \geq 0, ~ j = 1, \ldots M $. In $ \mathbb{R}^2 $ the first orthant is the first quadrant, in $ \mathbb{R}^3 $ the first orthant is the first octant and so on.} of the objective space and its shape is quite simple: either a sphere, a curve or a simplex.

DTLZ1 is an M-objective problem with a simple linear Pareto front. As pointed out by the authors \cite{Deb2005DTLZ}, the only difficulty provided by this problem is the convergence towards the Pareto front. The search space contains ($11^k-1$) local Pareto-optimal fronts to attract the MOEAs. DTLZ2-DTLZ3 use a function based on spherical coordinate system to determine the position of the points in the objective space. For $x_i \in [0,1], ~ i = 1, \ldots, M-1$ the function corresponds to the surface of a sphere in the first orthant of space $\mathbb{R}^M$. Similarly, the geometric characteristics of this surface make the objectives conflicting and an adequate distribution of the vectors $\mathbf{x}_p$ guarantees a good distribution of the points in the Pareto front. In DTLZ4, $x_i$ is replaced by $x_i^{\alpha}$, where $\alpha$ is a bias parameter, in order to introduce a bias and make the spatial distribution of the points in the objective space harder; $\alpha = 100$ is suggested in \cite{Deb2005DTLZ}. In DTLZ5 and DTLZ6, a slight modification in the auxiliary function turns the Pareto front into a curve contained in a sphere in the first orthant of the objective space in problems with three objectives. 
DTLZ7 to DTLZ9 problems do not use the spherical coordinate system in the $M$-dimensional space. The DTLZ7 presents a simple formulation for objectives $1$ to $M-1: f_i (\mathbf{x}) = x_i$ for $1 \geq i \geq M-1$. The last objective $f_M (\mathbf{x})$ is the only one dependent on the other variables of the problem. DTLZ7 presents $2^{M-1}$ disconnected regions in the Pareto front.  DTLZ8 and DTLZ9 are the only problems that present inequality constraints in this family. The former presents $M$ constraints and the latter $M-1$. The Pareto front of the DTLZ8 problem  with three objectives is composed of a straight segment and a triangular shaped flat surface, and the Pareto front of the DTLZ9 problem is quite similar to that presented by the DTLZ5 problem.

In the problems DTLZ1 to DTLZ7, the decision space has $N = M-1 + k$ variables. The first $M-1$ variables give the spatial location of the points in the objective space, while the remaining $k = N - M + 1$ variables are responsible for the convergence of points to the Pareto front. Thus, in these problems a vector in the decision space can be written as $\mathbf{x} = (\mathbf{x}_p, \mathbf{x}_d)$, where $\mathbf{x}_p$ is the portion responsible for the spatial location of the points in the objective space and $\mathbf{x}_d$ responsible for convergence. In DTLZ8 and DTLZ9, it is suggested $N = 10M$ variables \cite{Deb2005DTLZ}. The $k$ values suggested are $k = 5$ for DTLZ1, $k = 10$ for DTLZ2-DTLZ6 and $k = 20$ for DTLZ7. The optimal Pareto set for problems DTLZ1-DTLZ7 is $x_i \in [0,1]$ for $x_i \in \mathbf {x}_p$; $x_j = 0.5$ for $x_j \in \mathbf {x}_d$ in DTLZ1-DTLZ5; and $x_j = 0$ for $x_j \in \mathbf {x}_d $ in DTLZ7. Lastly, the solution for the DTLZ8 and DTLZ9 problems is not presented.

%%=========================================
\subsection{WFG Benchmark}
\label{subsec_wfg}

\citet{huband2006review} divided the desirable characteristics into those related to the fitness landscape and the Pareto optimal front geometry. Also, they analyzed the scalability and the separability of the MOO problems. According to the authors, the problem should be well defined for any number of objectives and be scalable, since a problem with more decision variables than objectives in general presents more difficulties to the optimizer. For a vector $\mathbf{x} = (x_1, \ldots, x_N) $ in the decision space, any variable $ x_i $ is classified in two ways: $ x_i $ is a distance parameter if its variation produces a new $ {\mathbf{y}} $ that changes the dominance relationship between $ F (\mathbf{y}) $ and $ F (\mathbf{x}) $. Otherwise, $ x_i $ is a position parameter. If a variable $ x_i $ has the same optimal value $ x_i^{\star} $ regardless of the values of the other decision variables, then this variable is separable. Otherwise, $ x_i $ is non-separable. If every variable of an objective $ f_i (\mathbf {x}) $ is separable, then this objective is separable. Consequently, if all objectives are separable, then the problem $ F (\mathbf {x}) $ is separable.
The solutions $ \mathbf{x}^{\star} = (x^{\star}_1, \ldots, x^{\star}_N) $ of the problem in the decision space are classified according to the location of $ x^{\star}_i $ in the interval $ [L_i, U_i ] $ where this variable is defined. If $ x^{\star}_i $ is close to the extremes $ L_i $ or $ U_i $, then $x^{\star}_i$  is on extremal parameter. Otherwise, if $ x^{\star}_i $ is located near the center of the interval $ [L_i, U_i] $, then $ x^{\star}_i $ is a central (medial) parameter.

With reference to the convergence in the objective space, the authors pointed out that a problem can be classified as unimodal or multimodal, where the deceptiveness is an specific type of multimodality. In a deceptive problem, the sub-optimal solutions lead the population of the MOEA to a region far from the one where the global optimum is located. A final aspect is the Pareto front. In a problem with $ M $ objectives, the Pareto front is, in general, a surface $S$ of dimension $ M-1 $. If the Pareto front dimension is less than $ M-1 $, then the problem is degenerate. This surface can be concave, convex, flat or a mixture of these formats. This surface can also be connected, i.e. given any two points $ A $ and $ B $ in $ S $, there is always a path $ c $ contained in $ S $ connecting these points. Otherwise, the Pareto front is disconnected. 

\citet{huband2006review} also indicated several recommendations for the construction of test problems, such as: a) extremal or central parameters should not be used; b) adjustable dimension of the decision and objective space; c) the search and objective space must be dissimilar, i.e., the variables in the search space with intervals of different sizes (dissimilar parameter domains), as well as the solutions in the objective space (dissimilar tradeoff ranges); d) the optimal solution should be known; and e) must present different shapes of Pareto front.

Based on these recommendations, the authors presented a nine functions toolkit, WFG1-WFG9. Starting from a vector of parameters $ \mathbf{z} $, a sequence of transformations is applied in order to obtain another vector $\mathbf{x} $ that adds the desired characteristics. The problem is then defined by minimizing the objectives $ f_i(\mathbf{x}), ~ 1 \leq i \leq M $. The vector $ \mathbf{z} $ has $ k + l = N \geq M $ positions, with the first $ k $ variables determining the position of $ F(\mathbf{x}) $ in the objective space and the last $ l $ variables responsible for the distance from $ F(\mathbf{x}) $ to the Pareto front of the problem. The transformed vector $ \mathbf{x} $ has $ M $ positions, the first $ M-1 $ coordinates being responsible for the position of $ F(\mathbf{x}) $ in the objective space and the last variable responsible for the distance from $ F(\mathbf{x}) $ to the Pareto front of the problem. In this way, this class of problems can be represented by a sequence of applications $\mathbf{Z} \xrightarrow{t(\mathbf{z})} \mathbf{X} \xrightarrow{F(\mathbf{x})} \mathbf{Y}$, where $\mathbf{Z}$ is the decision space, $ \mathbf{X} $ is the space of the parameters and $\mathbf{Y} $ is the objective space, with $ 0 \leq z_i \leq z_{i, max}$, and $ 0 \leq x_i \leq 1 $.

%%=========================================
\subsection{Other Test Suites}
\label{subsec_recent}

Since benchmark test problems are of great significance for the development of MOO algorithms, new test functions have been created to introduce new features and difficulties to compare the ability of the several MOEAs. 

More recently, \citet{wang2018generator} proposed a test problem generator that enables the design of MOO problems with complex Pareto front boundaries. The generator allows the researcher to control the feature of boundaries, consequently varying the difficulty for the MOEAs in achieving uniformity-diversity and breadth-diversity. \citet{matsumoto2019multiobjective} examined the influence of the shapes of the Pareto Front as well as the shape of the feasible region. Since scalability is not enough to impose difficulties for the MOEAs, the authors proposed a set of seven test problems with hexagon and triangular types of Pareto fronts. The results indicated differences between the algorithms used, given the different curvatures of the functions. \citet{yue2019novel} proposed a novel family with 12 scalable multimodal MOO problems with different characteristics, such as scalability, presence of local Pareto optimal solutions, non-uniformly distributed Pareto shapes and discrete Pareto front, being all of them continuous optimization problems. \citet{helbig2014benchmarks} and \citet{jiang2019scalable} focused on dynamic MOO (DMOO) problems. The former described a set of characteristics of an ideal set of DMOO benchmarking functions and proposed different problems for each characteristic. The latter proposed 15 scalable problems challenging the current dynamic algorithms to solve them. Ma and Wang \cite{mawang2019evolutionary} designed a test suite consisting of 14 problem instances for constrained multi-objective optimization, which tries to model characteristics extracted from real-world applications. Yu, Ji and Nolhofer \cite{yujinolhofer2019benchmark}, in turn, proposed a set of test problems whose Pareto fronts consist of complex knee regions, i.e. an important geometric feature on the Pareto-optimal front, ``where it requires an unfavorably large sacrifice in one objective to gain a small amount in other objectives'' \cite{yujinolhofer2019benchmark}. \citet{weise2018difficult} proposed a benchmark suite tunable towards different difficulty features for bit string based problems. Authough that work is not applicable to MaOPs and is from the discrete domain, it shows that the proposed idea of a tunable benchmark suite is interesting in optimization in general.

These recent works bring important contributions for the research and development of optimization algorithms in different ways, specially for modeling  features from practical problems and adding new aspects to the performance evaluation of  MOEAs. However, at the same time, this brings a notable drawback to the researchers developing new methods: one has to work with 3 to 6 benchmarks from the literature in order to compare different competing MOEA.

%%=========================================
\subsection{Limitations}
\label{subsec_limitationstestfunctions}

The main limitation of the ZDT family is that the problems are restricted to two objectives. Nevertheless, because it presents a large number of variables responsible for population convergence and it is easy to implement, this set is still valid to evaluate algorithms for problems with two objectives.

In the DTLZ family, the most used problems are DTLZ1 to DTLZ4. They are scalable to any number of objectives but have a small variety of Pareto Front shapes, namely a plane for DTLZ1 and a sphere for the others. In addition, the number of variables related to the positioning of solutions $ F (\mathbf{x}) $ in the objective space is restricted to $ M-1 $. The authors suggest a way to increase the difficulty of the problems by replacing the nominal value of the variable $ x_i $ ($ 1 \leq i \leq  M-1 $) by the meta-variable $ y_i $ given by the mean value of $ q $ variables, using Eq. \eqref{eq:Deb_mean}

\begin{equation}\label{eq:Deb_mean}
y_i=\frac{1}{q}\sum_{k=(i-1)q+1}^{iq}x_k
\end{equation}
which makes the spatial location of the points in the objective space depend on a larger number of variables. However, this strategy is not challenging enough to the current MOEAs. 

%The convergence is controlled by the $ g (\mathbf{x}) $ function defined by Eq. \eqref{eq:g13} for problems DTLZ1 and DTLZ3 and Eq. \eqref{eq:g24} for DTLZ4. Figure \ref{fig:g(x)} represents the shape of the functions $ g (\mathbf{x}) $ restricted to a single variable $ \mathbf{x} $ for DTLZ1 and DTLZ3 (Fig. \ref{fig:g_13}) and DTLZ2 and DTLZ4 (Fig. \ref{fig:g_24}). The Eq. \eqref{eq:g13} used for  DTLZ1 and DTLZ3 presents a multimodal problem, whereas the equation \eqref{eq:g24} makes DTLZ2 and DTLZ4 unimodal. 
DTLZ5 and DTLZ6 have a degenerate Pareto front, but present inconsistencies in problems with more than three objectives \cite{Huband2006}. DTLZ7 has a very simple formulation, $ f_i (\mathbf{x} ) = x_i, ~ 1 \leq i \leq M-1 $, with $ f_M (\mathbf{x}) $ the only objective with more elaborate algebraic expression. In addition, it has as an optimal solution an extremal value ($ x_i = 0, M \leq i \leq N $). The DTLZ8 and DTLZ9 problems are partially degenerate and their results are difficult to interpret in high dimensions, besides not presenting a known solution set for validation of results.

Another limitation also present in the DTLZ family is the lack of formulations related to robust optimization, as well as the absence of inequality and equality constraints. DTLZ8 and DTLZ9 problems, as mentioned, present inequality constraints, but without any possibility of customization. %Another relevant question, which is a consequence of the complexity of these functions, is its implementation and interpretation, which is not as simple as the other functions analyzed.

The WFG toolkit, in its turn, was carefully crafted, marked by the list of virtues and failures raised after extensive and rigorous analysis of the work presented so far. Although WFG offers a number of advantages, it also has some significant limitations. For example, a characteristic present in WFG1 is the idea of flat regions in the objective space, i.e. a region where small perturbations in the variables do not affect in a straightforward way the value of the objectives. A global optimum in a flat region may lead to similar results by MOEAs with different performance. The lack of geometric meaning of the transformations also makes it difficult to analyze the results obtained. The analysis of the results is restricted to the existing evaluation metrics.

The characteristics of the most classical test suites are also discussed in a number of works in the literature, see \citet{ishibuchi2016paretofronts}, \citet{huband2006review} and \citet{zapotecas2019review}. It is important to realize that even for the most popular test suites, such as ZDT, DTLZ and WFG, the well-diversified approximate solutions can also be easily attained by MOEAs. The most recent research has focused either on specific types of problems, as DMOO \cite{jiang2019scalable,yujinolhofer2019benchmark} or on variations on the Pareto Front shape \cite{matsumoto2019multiobjective}. However, the problems remain fixed and without any customization and the properties of deceptiveness and robustness are not included.

%%=========================================
%%=========================================
\section{A Customizable Family of Benchmark Problems}
\label{sec_newfamiltybenchproblems}

%%=========================================
\subsection{Desirable Characteristics}
\label{sub_desirablecharacteristics}

In principle, test functions should consist of a variety of problems that are capable to capture a wide variety of desirable characteristics. In order to list the main characteristics, we consider some recommendations that have already been made in the literature combined with our own ideas in Table \ref{tab_desiredcharacteristics}. The related references are also indicated.

\begin{table}
\caption{Listing of desirable features for test problems}
\label{tab_desiredcharacteristics}
\begin{tabular}{|p{3cm}|p{10cm}|p{2.5cm}|}
\hline
\textbf{Characteristic}  & \textbf{Description}& \textbf{Reference}  \\ \hline
Pareto optimal set known & Many performance measures require knowledge of the Pareto front. Additionally, the true Pareto front should be simple to generate and the spatial location of solutions in the objective space should be also known.  & \cite{ishibuchi2008reviewmaop}, \cite{huband2006review}, \cite{weise2018difficult}, \cite{zhang2007moead}    
\\ \hline
Scalability     & The test suite should be scalable for any number of objectives with free definition of number of variables as well as free definition of number of objectives.& \cite{maltese2018scalability}, \cite{ishibuchi2008reviewmaop}, \cite{wang2018generator}, \cite{Deb2005DTLZ}, \cite{huband2006review}, \cite{helbig2014benchmarks} 
\\ \hline
Pareto front geometries  & The Pareto front geometries include linear, nonlinear, convex, concave, mixed, discontinuous/disconnected, inverted and so on. As explained by \citet{yue2019novel}, generally a nonlinear geometry is harder than the linear one. Also, the discontinuous/disconnected may cause some algorithms to fail. Rich variety of shapes for the Pareto front is highly desirable. & \cite{wang2018generator}, \cite{matsumoto2019multiobjective}, \cite{yue2019novel}, \cite{helbig2014benchmarks} 
\\ \hline
Pareto set geometries    & The Pareto set geometries should include linear and nonlinear, connected and disconnected, and symmetric and non-symmetric. Also, the main idea is to evaluate the algorithms. & \cite{yue2019novel} 
\\ \hline
Constraints &   Set of inequality and equality constraints which are easy to interpret and identify their validity and violation.  & \cite{mawang2019evolutionary} \\ \hline
Modality    &   The function can be unimodal, multimodal or deceptive. A deceptive problem may cause the most difficulty for EAs, swarm-based algorithms and other meta-heuristics, since the sub-optimal solutions lead the population to a region far from the one where the global optimum is located.    & \cite{zapotecas2019review}, \cite{helbig2014benchmarks}, \cite{Huband2006} 
\\ \hline
Separability    & There are two concepts of separability involved: one is related to the separability between distance and position parameters. The other is related to controlling the degree of separability of the function in terms of the optimal values.  & \cite{maltese2018scalability}, \cite{zapotecas2019review}, \cite{yujinolhofer2019benchmark} 
\\ \hline
Dissimilarity   & The parameters of the test problem and the tradeoff ranges in each objective should have domains of dissimilar magnitude.    & \cite{cheng2017benchmark,Huband2006}
\\ \hline
Independent parameters   & Parameters in each function that independently adjust the challenges presented to the DM regarding to the convergence and coverage. %Podem ser de posição, distância e mixto by Huband2006
& \cite{wang2018generator,zapotecas2019review}
\\ \hline
No extremal or medial parameters &  Both are to prevent exploitation by truncation, based on correction operators in the case of extremal parameters and on intermediate recombination in the case of medial parameters.  & \cite{huband2006review} 
\\ \hline
Bias    &   It represents the difficulty in sampling parts of the Pareto front causing a natural impact on the search process.    & \cite{wang2018generator}, \cite{huband2006review}, \cite{matsumoto2019multiobjective}, \cite{li2016biased} 
\\ \hline
Multi-modality with brittle global optima and robust local optima & It is possible to formulate a problem where each objective presents a global optimum and several closer sub-optimal solutions. However, in only one of the sub-optimal solutions, robustness is observed. The evaluation of this robust sub-optimal solution is stable, that is, the objectives present a minimum variation for a significant range of values in the decision space. A robust solution should still work satisfactorily when the design variables change slightly. This feature is desirable in many practical problems, since the presence of noise, disturbances and variability is always frequent. & \cite{yue2019novel}, \cite{Yue2018}, \cite{Liang2019}, \cite{Preuss2006}, \cite{Jin2005}
\\ \hline
Customization    &   Benchmark functions should have parameters to control specific features of the problem. The user should be able to generate different instances with controllable difficulties.  &  \cite{wang2018generator, weise2018difficult}
\\ \hline
\end{tabular}
\end{table}

Trying to cover these directives, this section introduces a new set of scalable and customizable benchmark problems for MaOPs. The proposed test suite uses the bottom-up approach \cite{Deb2005DTLZ,zapotecas2019review}. Once the Pareto optimal front, the objective space, and the decision space are separately constructed, this method has facilitated the design of MOO problems. We propose the minimization problem described in Eq. \eqref{eq:Fbase}, using either a multiplicative or an additive approach.

\begin{align} \label{eq:Fbase}
\min F(\mathbf{x}), ~& F: \mathbb{R}^N  \to \mathbb{R}^M \nonumber\\  
\mbox{subject to } &
\begin{cases}
\phi_i(\mathbf{x})\leq A_i, & i=1 \ldots k_1 \\
\xi_j(\mathbf{x}) = I_j, & j=1
\ldots k_2, ~ I_j\in \{1 \ldots M\}
\end{cases}
\end{align}

\noindent where $F(\mathbf{x}) = F_p(\mathbf{x})F_d(\mathbf{x})$ or $F(\mathbf{x}) = F_p(\mathbf{x})+ F_d(\mathbf{x})$. In this problem, a vector $\mathbf{x}$ in decision space $\mathbb{R}^N$ can be written as $\mathbf{x}=(\mathbf{x}_p,\mathbf{x}_d) \in \mathbb{R}^N = \mathbb{R}^R \times \mathbb{R}^S$, where $\mathbf{x}_p=(x_1, \ldots, x_R) \in \mathbb{R}^R$ and $-1 \leq x_i \leq 1$,  $i=1 \ldots R$, is a vector with $R$ coordinates, responsible for the positioning of points $F (\mathbf{x})$ in the objective space and $\mathbf{x}_d=(x_1, \ldots, x_S) \in \mathbb{R}^S$ with $0 \leq x_j \leq 1$,  $j=1 \ldots S$, is a vector with $ S $ coordinates, responsible for the convergence of $F (\mathbf{x})$. $\phi_i(\mathbf{x})$ and $\xi_j(\mathbf{x})$ are constrains. The details of the functions $F_p (\mathbf{x})$ and $F_d (\mathbf{x})$ are discussed next.

%%=========================================
%\subsection{Function \texorpdfstring{$F_p(\mathbf{x})$}}
\subsection{Function {$F_p(\mathbf{x})$}}
\label{subsec_fpx}

The function $F_p (\mathbf{x})$ is responsible for the relative position of the points in the objective space and the conflict among the objectives. The basic definition is given in Eq. \eqref{eq:T}, which describes the surface of a hypersphere of radius $1$ in the first orthant of space $\mathbb{R}^ M$ in spherical coordinates and depends on at least $M- 1$ variables. The preference for this equation is due to the fact that it describes the spatial location with simplicity and precision. Two modifications are presented that aim to increase the number of variables and the diversity of formats to this surface. 

One way to increase the difficulty presented to the optimizer in finding the best distribution of points in objective space is to replace $x_i$ of $F_p (\mathbf{x})$ by the meta-variable $y_i$ defined in Eq. \eqref{eq:meta_variavel}. It corresponds to the mean of $q + t$ values in the set of variables $x_{i(i-1) q + 1}, \ldots, x_{iq + t}$ from the original decision space.

\begin{equation}\label{eq:meta_variavel}
y_i= \frac{1}{q+t}\left|\sum_{j=(i-1)q+1}^{iq+t} x_j\right| , -1 \leq x_j \leq 1
\end{equation}

Note that the difference between the meta-variables presented in Eqs. \eqref{eq:Deb_mean} and \eqref{eq:meta_variavel} is the sum of $t$ at the upper bound of the summation. Unlike the meta-variable \eqref{eq:Deb_mean} proposed for the DTLZ family, in Eq. \eqref{eq:meta_variavel}, each sub-interval where it is calculated the average, shares $t$ elements with the previous sub-interval and $t$ elements with the next sub-interval, increasing the dependency between the decision space variables.
%Entretanto é necessário que 2t + 1 < q para que cada sub-intervalo apresente pelo menos uma variável xi independente dos intervalos vizinhos de modo que exista pelo menos uma solução que alcance qualquer região da Fronteira Pareto.
However, it is necessary that $2t + 1 < q$ for each sub-interval, guaranteeing the presence of at least one independent variable $x_i$ from neighbouring intervals, and, consequently, that there is at least one solution that reaches any region of the Pareto front. Thus, the subspace $\mathbb{R}^R$ which is responsible for positioning the points in decision space will have $(M-1) q + t$ variables. The distribution of the meta-variables $\mathbf{y}_i $ in the parameter vector $ \mathbf{x} $ is presented in Eq. \eqref{eq:overlap}.

\begin{equation}\label{eq:overlap}
\begin{split}
\mathbf{x}=&(\UOLoverbrace{x_1, \ldots,}[x_{q+1}, \ldots, x_{q+t},]^{y_1}
\UOLunderbrace{x_{q+t+1},\ldots,x_{2q},}[ x_{2q+1},\ldots, x_{2q+t},]_{y_2}
\UOLoverbrace{x_{2q+t+1}\ldots, \ldots, x_{3q+t},}^{y_3}, \ldots \\ 
& \ldots ,\overbrace{x_{(M-2)q+t+1},\ldots,x_{(M-1)q+t}}^{y_{M-1}} ) \\ 
y_1=& \frac{1}{q+t}\left|\sum_{j=1}^{q+t} x_j\right|, \\ 
y_2=& \frac{1}{q+t}\left|\sum_{j=q+1}^{2q+t} x_j\right|,\\ 
&\vdots \\
y_{M-1}=& \frac{1}{q+t}\left|\sum_{j=(M-2)q+1}^{(M-1)q+t} x_j\right| 
\end{split}
\end{equation}

\begin{figure}[htb]	
	\centering
	\subfloat[Meta-variable $y$ with $q=10$ and $t=0$. \label{fig:dtlz2_s0}]{\includegraphics[width=.5\linewidth]{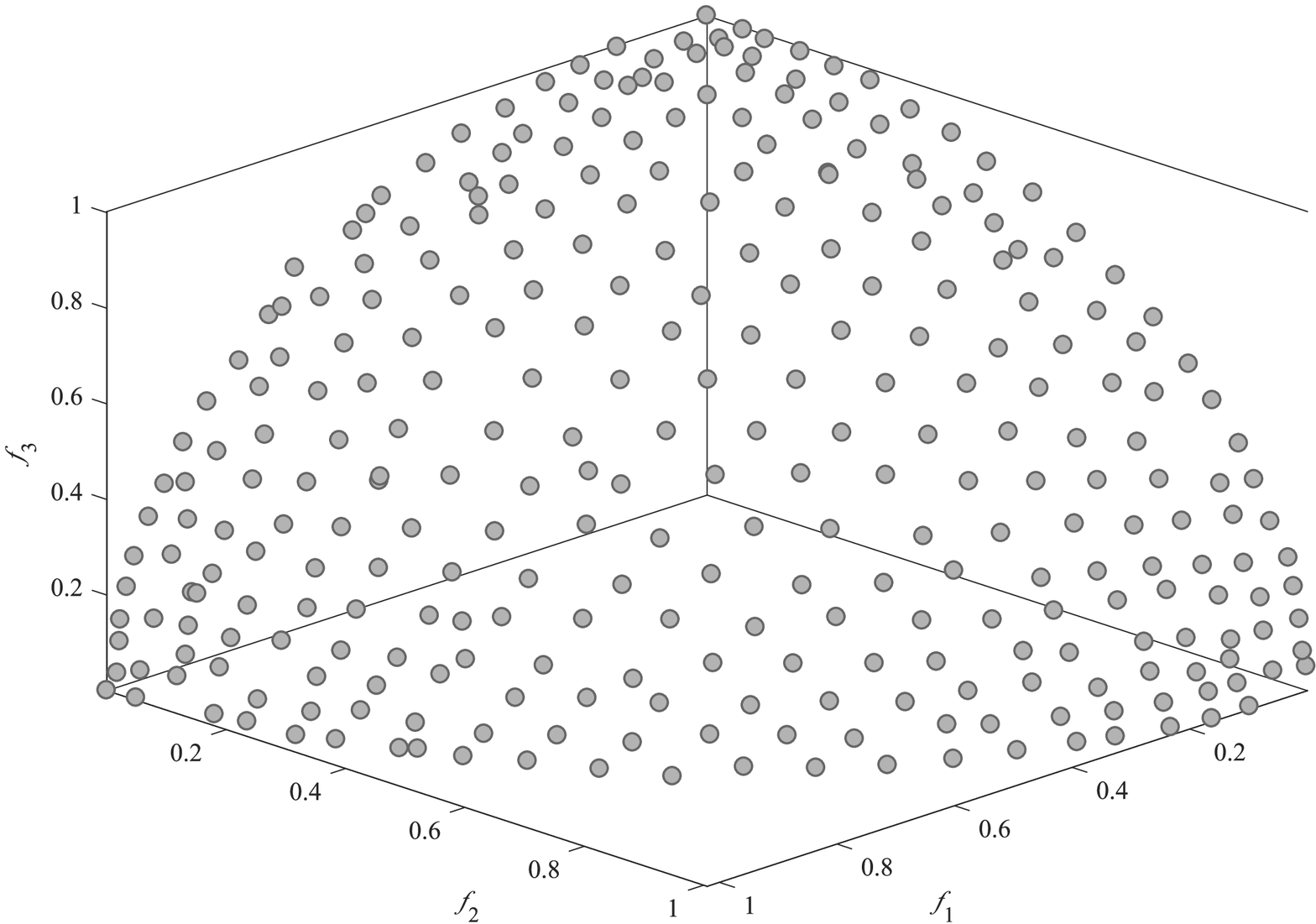}}
	\subfloat[Meta-variable $y$ with $q=10$ and $t=4$. \label{fig:dtlz2_s4}]{\includegraphics[width=.5\linewidth]{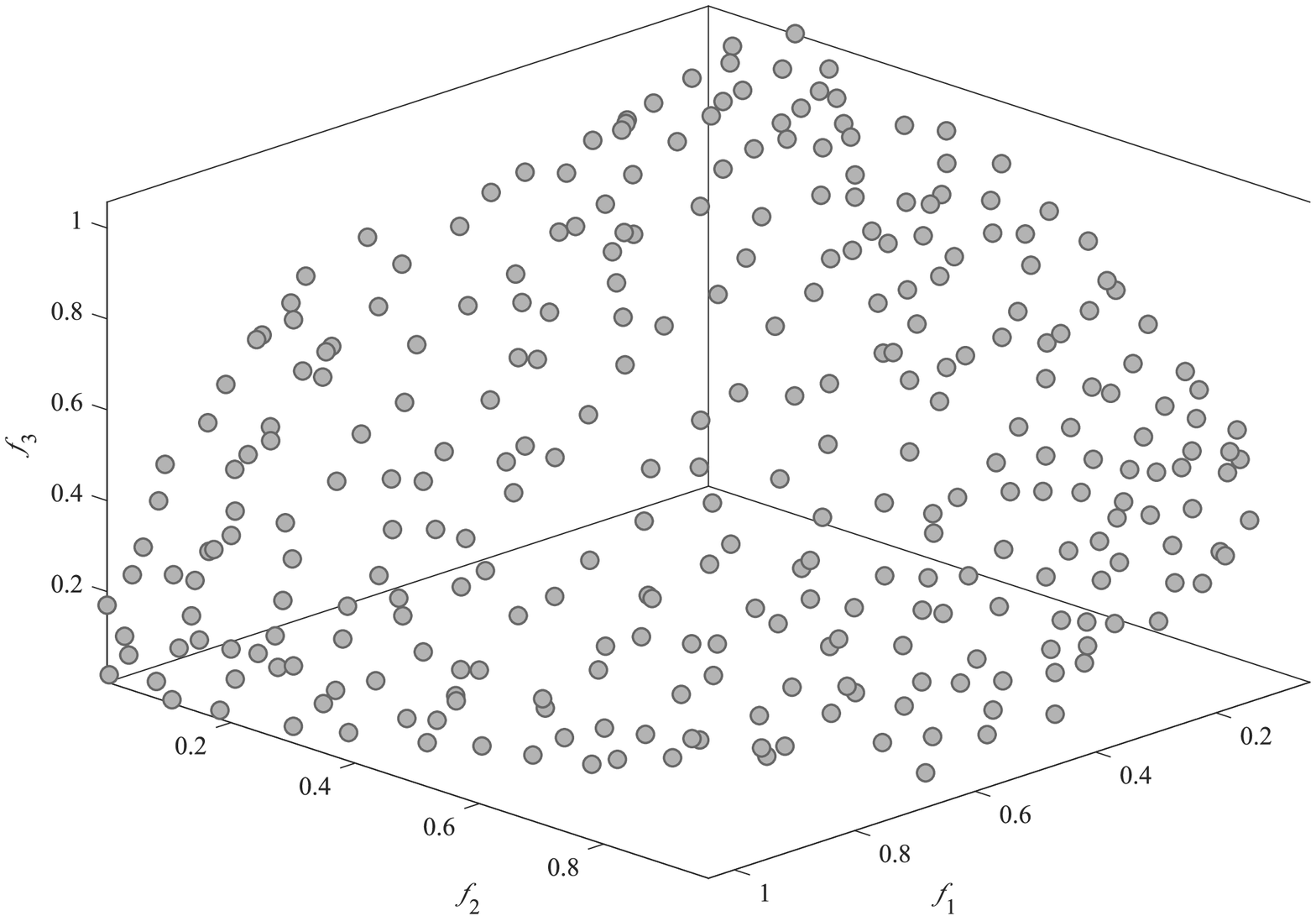}}
	\caption{Implications of the parameter $t$ in the meta-variable $y$ applied in DTLZ2 problem}
	\label{fig:s0_X_s4}
\end{figure}

Figure \ref{fig:s0_X_s4} illustrates the effect of using the meta-variables $\mathbf{y}_i$ on a MOO problem. The DTLZ2 problem with three objectives was selected and it was used the variable $\mathbf{y}$ in Eq. \eqref{eq:meta_variavel} with parameter $q = 10$.  In Figure \ref{fig:dtlz2_s0} the obtained solution with $t = 0$ is presented, which corresponds to the use of the variable $\mathbf{y}$ defined by Eq. \eqref{eq:Deb_mean}, whereas in Figure \ref{fig:dtlz2_s4} the parameter $t = 4$ was used. For both, the NSGA-III algorithm \cite{deb2014nsgaiii} available in the platEMO platform \cite{tian2017platemo} was used with the same hyper-parameters. When comparing Figures \ref{fig:dtlz2_s0} and \ref{fig:dtlz2_s4}, we immediately notice the degradation effect of the spatial distribution over the solutions by using the proposed meta-variables.

The use of this meta-variable has two important advantages:
\begin{enumerate}
    \item Considering the decomposition of vector $\mathbf{x}=(\mathbf{x}_p,\mathbf{x}_d)$ in decision space the $\mathbf{x}_p$ component generally has $M-1$ components. Thus, the increase in the number of decision space variables corresponds to the increase in the number of variables of the $\mathbf{x}_d$ component. The meta-variable allows scaling of the $\mathbf{x}_p$ vector to more than $M-1$ variables. This more flexible form of design space scaling allows the use of this class of test problems in large scale MOO algorithms \cite{7553457}.
    \item In general, the vector $\mathbf{x}_p$ is randomly initialized according to a uniform distribution over a range $[a, b]$. This distribution has mean $(b-a) / 2$ and variance $(b-a)^2/12$, making the initial population symmetrically distributed in the central region of this range. The meta-variable is initialized in the same way. However, after its transformation, the initial population accumulates non-symmetrically at the beginning of the interval. This new biased configuration poses an unprecedented challenge to optimization algorithms.
\end{enumerate}.

\begin{figure}[htb]
	\centering
	\includegraphics[width=0.7\linewidth]{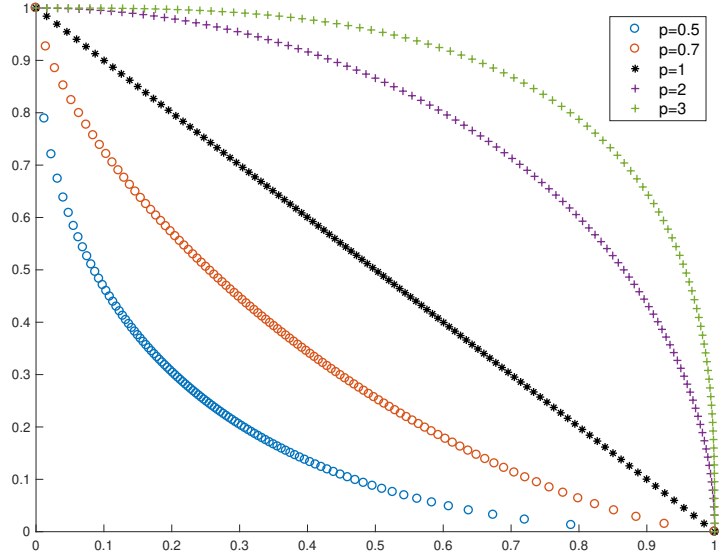}
	\caption{Points with different $p-$norm value in $\mathbb{R}^2$ space.}
	\label{fig:p-norm-2d}
\end{figure}

The shape of this surface can be controlled by changing the norm of points. If $\mathbf{x} = (x_1, \ldots, x_M)$ is a point in $\mathbb{R}^M$, a $p$ -norm $ || \mathbf{x} ||_p $ (also called $\ell_p$ norm) of $\mathbf{x}$ is given by the Eq. \eqref{eq:p-norm}

\begin{equation}\label{eq:p-norm}
||\mathbf{x}||_p = \left( \sum_{i=1}^{M} |x_i|^p \right)^{1/p}
\end{equation}
where $p \geq 1$ and $| x_i |$ is the absolute value of $x_i$ (note that the absolute value of $ x_i $ is the norm $ \ell_2 $ of $ x_i $, i.e., $ | x_i | = || x_i || _2 $.). If $ p = 2 $ the $p$-norm is called the Euclidean norm and if $p = 1$ it is the Manhattan norm (or  taxicab norm). The infinite norm (or Tchebycheff norm) of the vector $ \mathbf{x} \in \mathbb{R}^M $, denoted by $ || \mathbf{x} ||_ {\infty}$, is defined as $||\mathbf{x}||_{\infty} = \max |x_i|, 1 \leq i \leq M$. So, consequently $\lim_{p \to \infty} ||\mathbf{x}||_p = ||\mathbf{x}||_{\infty}$. If $ 0 < p <1 $, the expression \eqref{eq:p-norm} does not define a norm, but a quasi-norm, since the triangular inequality is not satisfied \cite{Rudin1991}. Nonetheless, $ p $ values greater than zero are going to be used in Eq. \eqref{eq:p-norm} to produce constant norm (or quasi-norm) surfaces.  Figure \ref{fig:p-norm-2d} presents constant norm curves with different $ p $ values in $ \mathbb{R}^2 $ space.

In this way, the function $F_p(\mathbf{x})$ can be defined by:

\begin{equation}\label{eq:F_d}
F_p(\mathbf{x})=\frac{T(\mathbf{x})}{h(\mathbf{x})}
\end{equation}
where $ T (\mathbf{x}) $ is responsible for the distribution of the points in objective space, defined by Eq. \eqref{eq:T}, dependent on the meta-variable  $y_i= \dfrac{1}{q+t}\left|\sum_{j=(i-1)q+1}^{iq+t} x_j\right|$ and $h(\mathbf{x})=||T(\mathbf{x})||_p$.

\begin{equation}\label{eq:T}
T(\mathbf{x})=
\left\lbrace 
\begin{aligned}
t_1(\mathbf{x})	   & =\cos(y_1 \pi/2)\cos(y_2 \pi/2) \ldots \cos(y_{M-2}\pi/2)\cos(y_{M-1}\pi/2)\\
t_2(\mathbf{x})	   & =\cos(y_1 \pi/2)\cos(y_2 \pi/2) \ldots \cos(y_{M-2}\pi/2)\sin(y_{M-1}\pi/2)\\
t_3(\mathbf{x})	   & =\cos(y_1 \pi/2)\cos(y_2 \pi/2) \ldots \sin(y_{M-2}\pi/2)\\
\vdots			   & \\
t_{M-1}(\mathbf{x})& =\cos(y_1 \pi/2)\sin(y_2 \pi/2)\\
t_M(\mathbf{x})	   & =  \sin(y_1 \pi/2)	
\end{aligned} 
\right.
\end{equation}

Another desirable characteristic is the dissimilarity of the objectives. In general, each objective of benchmark problems for multiobjective optimization has its optimal values limited in the range $[0,1]$, i.e. $0 \leq f_i (\mathbf{x}) \leq 1$, but this hardly reflects real-world problems. Some authors, such as \citet{Huband2006} and \citet{cheng2017benchmark}, presented functions with dissimilar objectives, being $0 \leq f_i (\mathbf{x}) \leq \xi$, with $\xi$ a power or a multiple of 2. However, these problems only present objectives with non-negative values. An optimization problem with dissimilar objectives with positive and negative values is presented in Eq. \eqref{eq:D}

\begin{equation}\label{eq:D}
D(\mathbf{x})=
\left\lbrace 
\begin{aligned}
d_1(\mathbf{x})	   & =2(2f_1(\mathbf{x}) - 1)\\
d_2(\mathbf{x})	   & =4(2f_2(\mathbf{x}) - 1)\\
d_3(\mathbf{x})	   & =6(2f_3(\mathbf{x}) - 1)\\
\vdots			   & \\
d_M(\mathbf{x})	   & = (2M)(2f_M(\mathbf{x}) - 1)
\end{aligned} 
\right.
\end{equation}
where $f_i(\mathbf{x})$ represents the $i$-th objective obtained after the evaluation of $F(\mathbf{x})=F_p(\mathbf{x})F_d(\mathbf{x})$.

Once $0 \leq f_i (\mathbf{x}) \leq 1$, then $ -2i \leq d_i (\mathbf{x}) \leq 2i $. In fact, if $f_i(\mathbf{x})=0$ then $d_i(\mathbf{x})= 2i(0 - 1)=-2i$ as well as if $f_i(\mathbf{x})=1$ then $d_i(\mathbf{x})= 2i((2 - 1)=2i$.  Note that the function $D (\mathbf{x})$ transfers the origin from objective space $O = (0, \ldots, 0)$ to the point $O' = (-2, -4, \ldots, -2M)$, but it does not change the surface's  concavity.

%%=========================================
\subsection{Function $F_d(\mathbf{x})$}
\label{subsec_fdx}

This function is responsible for the convergence of points towards the Pareto front and, in some cases, for the shape of the Pareto front. This function makes use of the auxiliary functions $g (\mathbf{x})$ and $\phi(\mathbf{x})$.

%%=========================================
\subsubsection{Auxiliary Function $\phi(\mathbf{x})$}
\label{subsubsec_phix}

The auxiliary function $\phi(\mathbf{x})$ aims to incorporate some information about the position of $F(\mathbf{x}) $ relative to the hyper-diagonal $\mathbf{d} = (1,1 , \ldots, 1)$ (or another vector $\mathbf{d}$) in the objective space. For the point $\mathbf{x} \in \mathbb{R}^N$ in the decision space, 
the point $F_p (\mathbf{x})$ is on a surface of $p$-norm (quasi-norm) constant in the objective space, with $|| F_p (\mathbf{x}) ||_p = 1 $. The $\varphi$ angle between the hyper-diagonal $ \mathbf{d} $ and $ F_p (\mathbf{x}) $ is calculated by means of 

\begin{equation} \label{eq:angulo}
\varphi(\mathbf{x}) = \arccos\left( \frac{\mathbf{d} \cdot F_p(\mathbf{x})^T }{|\mathbf{d}| |F_p(\mathbf{x})|} \right)
\end{equation}

This value must be normalized into the range $ [0,1] $. For the vector $\mathbf{d} = (1,1 , \ldots, 1)$, the angle $\varphi$ is maximal if the vector defined by $ F_p (\mathbf{x}) $ is aligned with some vector of the canonical base $ \mathbf{e}_i = (0, \ldots, 1, \ldots, 0) $, that is, if $ F_p (\mathbf{x}) =\lambda \mathbf{e_i} $ for some $ \lambda > 0 $. In this case, the maximum angle is $\varphi_{\max}=\arccos\left( \frac{1}{\sqrt{M}}\right)$ and the value of the normalized distance function $\phi(\mathbf{x}_d)$ is given by
	
\begin{equation}\label{eq:phi}
\phi(\mathbf{x})=\frac{\varphi(\mathbf{x})}{\varphi_{\max}}   
\end{equation}

%%=========================================
\subsubsection{Auxiliary Function $g(\mathbf{x})$}
\label{subsubsec_gx}

The auxiliary function $ g (\mathbf{x}) $ is responsible for the convergence of the points in the objective space. In this work we present two versions: deceptive and multi-modal with brittle global optima and stable local optima. Other versions of this function can be incorporated into this proposal in order to satisfy some special need.
	
The first version, a parameterized deceptive function, is characterized by the presence of a global optimum and two local minima. Another relevant characteristic of this function is the influence of the relative position of the point in the objective space. The relative position is given by the function $\phi(x)$ (Eq. \eqref{eq:phi}). The topology of this problem favors the sub-optimal solutions, making the global optimum difficult to achieve. The deceptive function proposed, $g (\mathbf{x})$, is defined by Eq. \eqref{eq:g_deceptiva}. It presents, for each variable $x_i \in \mathbf{x}_d$, two local minima (in $x_i = 0$ and $x_i = 1 $) and one global (in $x_i=v$) optima located in a deep valley of width $2r$.

\begin{figure}[htb]
	\centering
	\subfloat[\label{fig:g_hole_def} Elements of the deceptive function $g(\mathbf{x})$.]{\includegraphics[width=.5\linewidth]{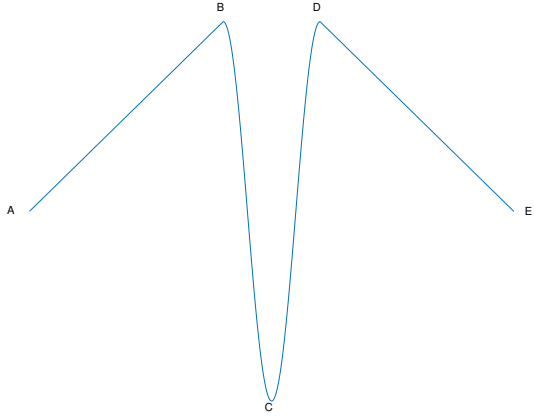}}
	\subfloat[\label{fig:g_decep} Function $g(\mathbf{x})$ with one variable $x_i$ and parameters $r=0.1,0.05$ and $0.01$.]{\includegraphics[width=.5\linewidth]{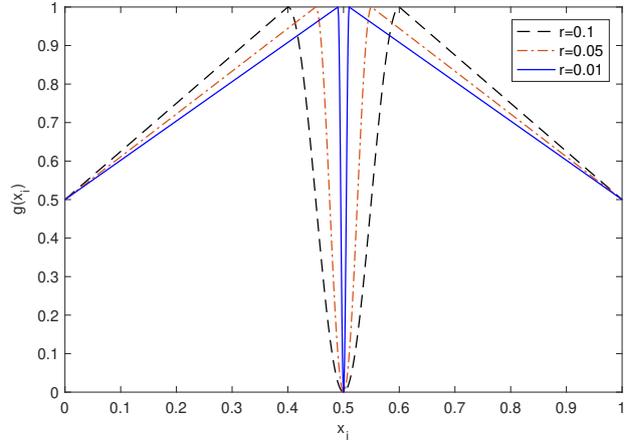}}\\
	\subfloat[\label{fig:g_hole_3D} $g(\mathbf{x})$ variation when $\phi$ ranges from 0 to 1. ]{\includegraphics[width=0.5\linewidth]{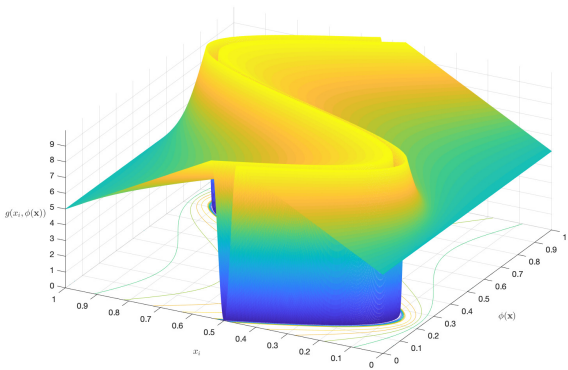}}
	\subfloat[\label{fig:g_hole_x1_k10} Contour lines of the deceptive function $g(\mathbf{x})$ for $i=1$ and $k=5$ for a single variable $\mathbf{x}$.
	]{\includegraphics[width=0.5\linewidth]{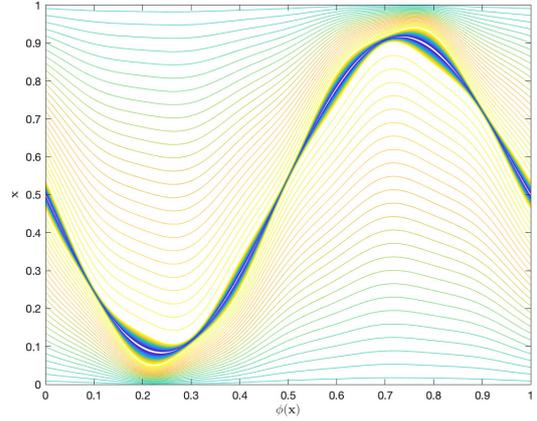}}
	\caption{Deceptive function $g(\mathbf{x})$}
	\label{fig:g_hole_01}
\end{figure}

The Figure \ref{fig:g_hole_def} presents the construction details of the function $g(\mathbf{x})$. If $ 0 \leq x_i \leq v-r $ (and $ v + r \leq x_i \leq 1 $), the function $ g (\mathbf{x}) $ is a line connecting the points $A$ and $B$ ($D$ and $E$ respectively). If $v-r < x_i < v + r$ then $g(\mathbf{x})$ is a complete cycle of the cosine trigonometric function connecting the points $B, ~ C$ and $D$. The $v$ and $r$ parameters do the correlation between the position and distance, respectively, of a point in the objective space and introduce a bias in the decision space. They are defined as follows:

\begin{align}
    v(\mathbf{x}_p, x_i)& =\frac{1.2 +\sin\left(2\pi(1-\phi(\mathbf{x}_p))^{1.05i}\right)}{2.4} \label{eq:v}\\
    r(x_i)& =0.015\cos(2k \pi\phi(\mathbf{x}_p)) + 0.025 \label{eq:r}
\end{align}

This special format has as motivation to hide the location of the optimal value $x_i$ by a small opening of the valley determined by the parameter $r$, leading the population to the local minima located at the beginning and the end of the range. Note that $r(x_i)$ and $v(x_i)$ use the normalized function $\phi(\mathbf{x})$. As seen before, this function estimates the relative position of a point in the objective space, controlling the range of the valley (segment $BD$ in the \ref{fig:g_hole_def}). In Eq. \eqref{eq:r}, the factor $k=1$ produces two large ranges and one narrow range. In a general way, $k+1$ large valleys and $k$ narrow valleys will be produced. Figure \ref{fig:g_hole_x1_k10} shows the contour lines of $g(\mathbf{x})$ for the first variable of the vector $\mathbf{x}$ using $k=5$.

The proposed deceptive function $g(\mathbf{x})$ is defined by

\begin{equation} \label{eq:g_deceptiva}
g(\mathbf{x})=\sum_{x_i \in \mathbf{x}_d} z(\mathbf{x}_p,x_i)
\end{equation}
where
\begin{align}\label{eq:hole_cases}
z(x_i) =
\left\lbrace 
\begin{array}{lcl}
\frac{5(x_i+r-v)}{v-r} + 10 & \mbox{ if } & 0 \leq  x_i < v - r \\
5(\cos(\frac{x_i+r-v}{r}\pi)+1) &\mbox{ if } &v-r \leq  x_i \leq v + r \\
\frac{5(x_i-v-r)}{v+r-1} & \mbox{ if } &v+r <  x_i \leq 1 \\
\end{array}
\right.
\end{align}

\begin{figure}[htb]
	\centering
	\subfloat[\label{fig:g_hole_x1}$i=1$]{\includegraphics[width=.5\linewidth]{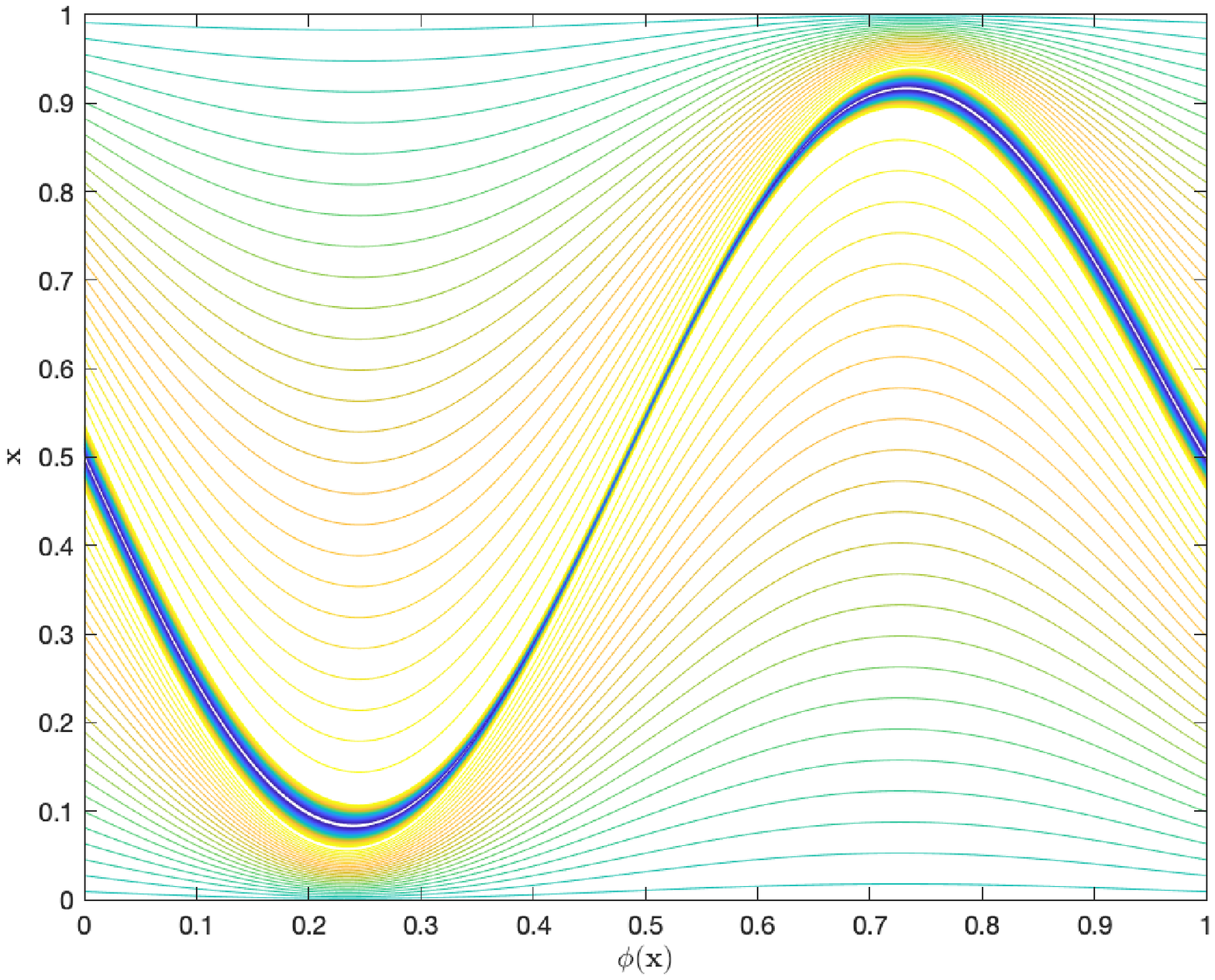}}
	\subfloat[\label{fig:g_hole_x5}$i=5$]{\includegraphics[width=.5\linewidth]{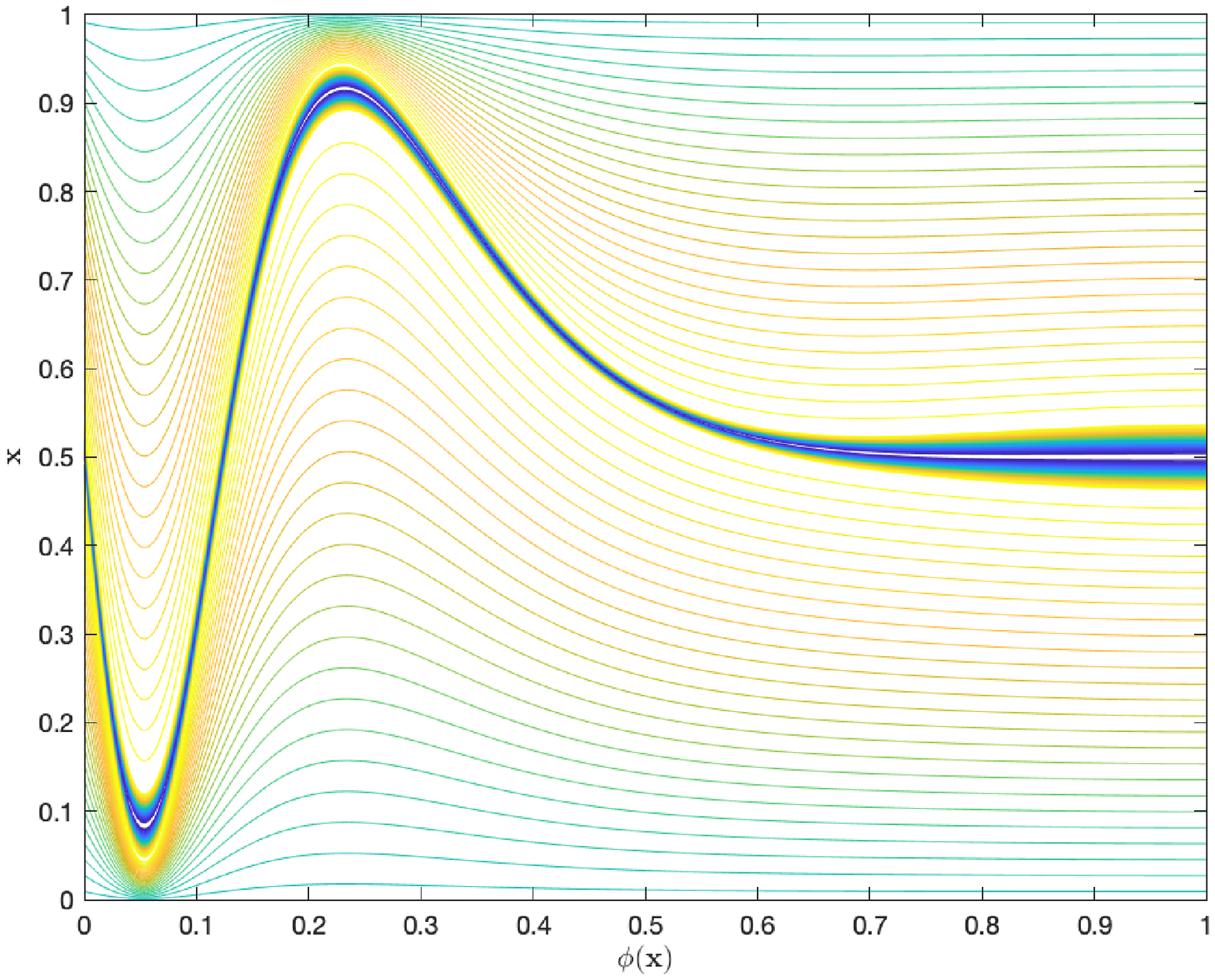}}\\
	\subfloat[\label{fig:g_hole_x10}$i=10$]{\includegraphics[width=.5\linewidth]{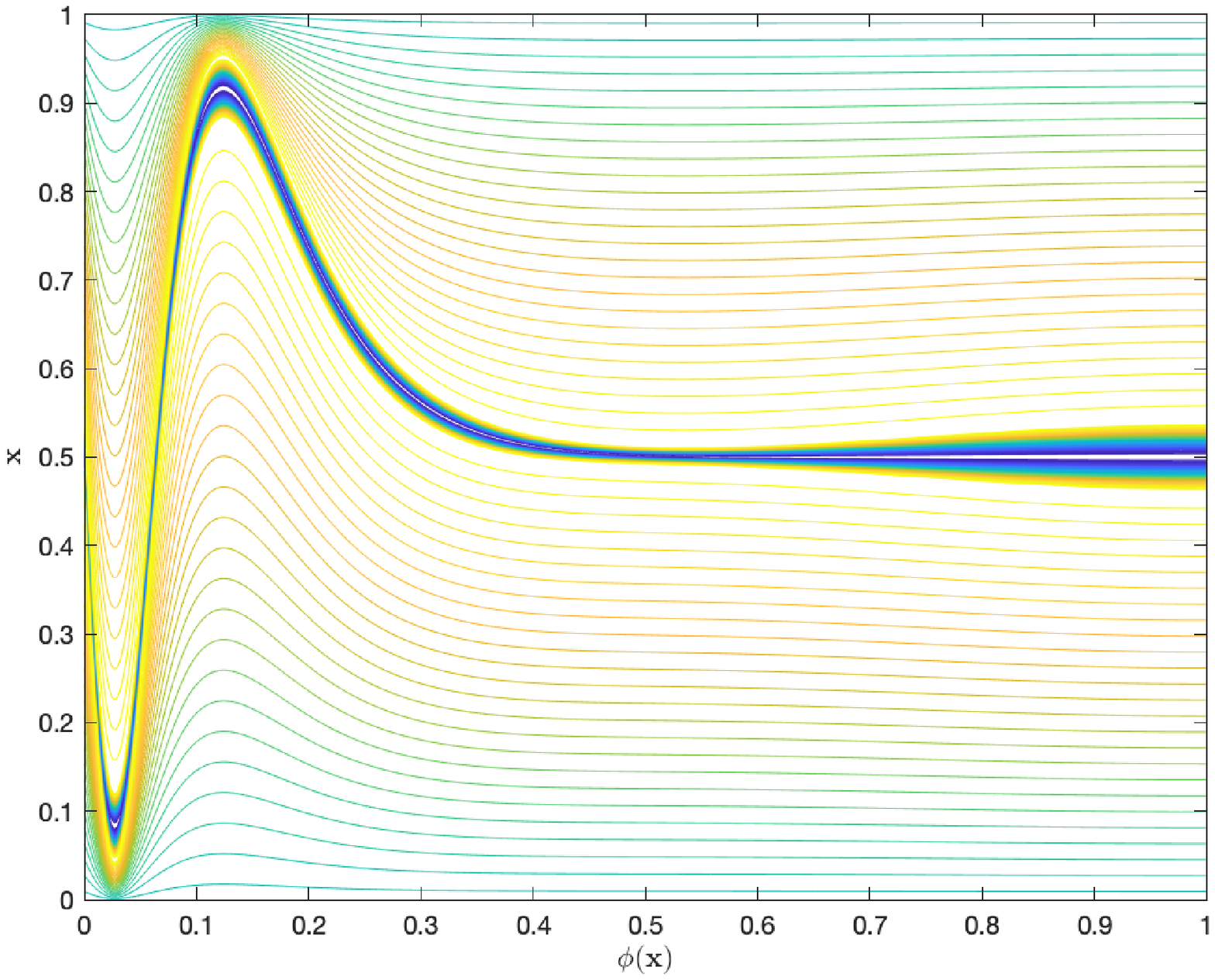}}
	\subfloat[\label{fig:g_hole_x20}$i=20$]{\includegraphics[width=.5\linewidth]{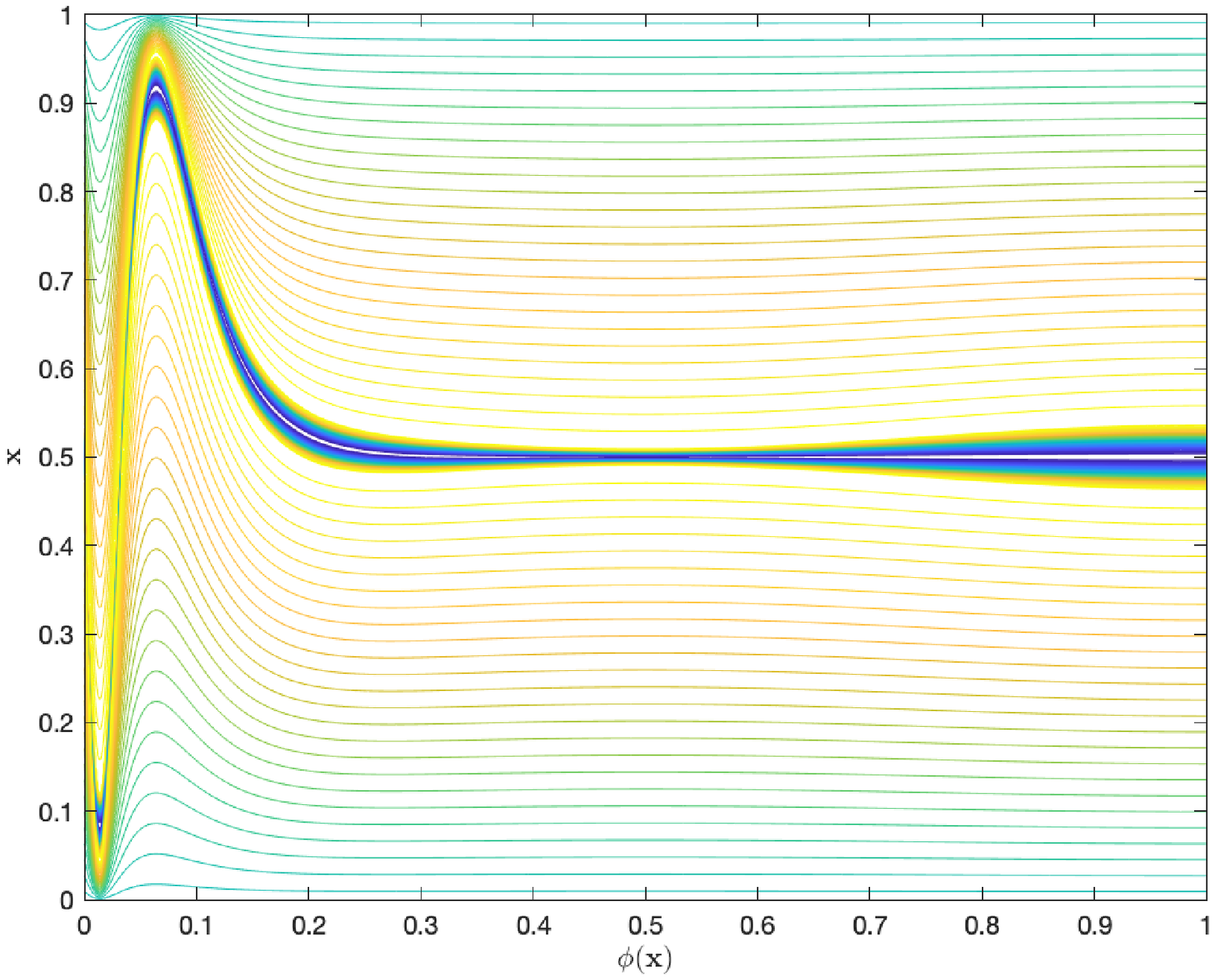}}
	\caption{Contour lines of the deceptive function $g(\mathbf{x})$ for $i=1,5,10,20$ and $k=1$ for a single variable $\mathbf{x}$}
		\label{fig:g_hole_02}
\end{figure}

Figure \ref{fig:g_decep} presents the function $g(\mathbf{x})$ for a single variable $x_i\in [0,1]$ with $r$ parameter equal to 0.1, 0.05 and 0.01. Figure \ref{fig:g_hole_3D} presents the variation of $g(\mathbf{x})$ when $\phi(\mathbf{x})$ and the first variable $x_1$  of $\mathbf{x}_d$ varies from 0 to 1. Figures \ref{fig:g_hole_x1} to \ref{fig:g_hole_x20} show the contour lines of $g(\mathbf{x})$ for $i=1,5,10$ and $20$, respectively.

The second version for the function $g(\mathbf{x})$ presents a special multi-modal distance function, with a brittle global optimum and local optima. The global optimum and some of the local optima near it are sensitive to the presence of noise on their optimal values in the Decision Space.  However, there is a region with stable local optima, whose values present little variation when this same noise is added to the optimal values in the Decision Space. This type of function is appropriate for the validation of algorithms designed for robust optimization \cite{Petersen2014, Meneghini2016, he2018robust,Wang_2018}, whose goal is to find stable solutions, that is, solutions that, when evaluated, show little variation when some noise is added in their neighborhood. The multi-modal function $g(\mathbf{x})$ is the combination of logistic and trigonometric functions. It has no extremal variables and the robust solutions are located in the $(0.1,0.3)$ range. The global optimal solution is $x_i = 0.60$ (precisely $x_i = 0.600066066066066 \ldots$) but it is sensitive to noise in its neighborhood. The robust $g(\mathbf {x})$ is defined by Eq. \eqref{eq:g_robusta}.

\begin{equation}\label{eq:g_robusta}
g(\mathbf{x})=\sum_{x_i \in \mathbf{x}_d}\left[-w(x_i)\left( y(x_i)-z(x_i)\right)+\frac{y(x_i) - 1}{2} + e^{-60 x_i} + 0.631\right]
\end{equation}
where
\begin{align}
y(x_i) \quad & =\frac{1}{1+ e^{-20(x_i - 0.6)}}\\
z(x_i) \quad & =\frac{1}{1+ e^{-20(x_i - 0.7)}}\\
w(x_i) \quad & =\cos(40 \pi x_i)
\end{align}

\begin{figure}[!htb]
	\centering
	\subfloat[\label{fig:fo_robust} Multi-modal function $g(\mathbf{x})$ for one variable $0 \leq x \leq 1$ .]{\includegraphics[width=0.5\linewidth]{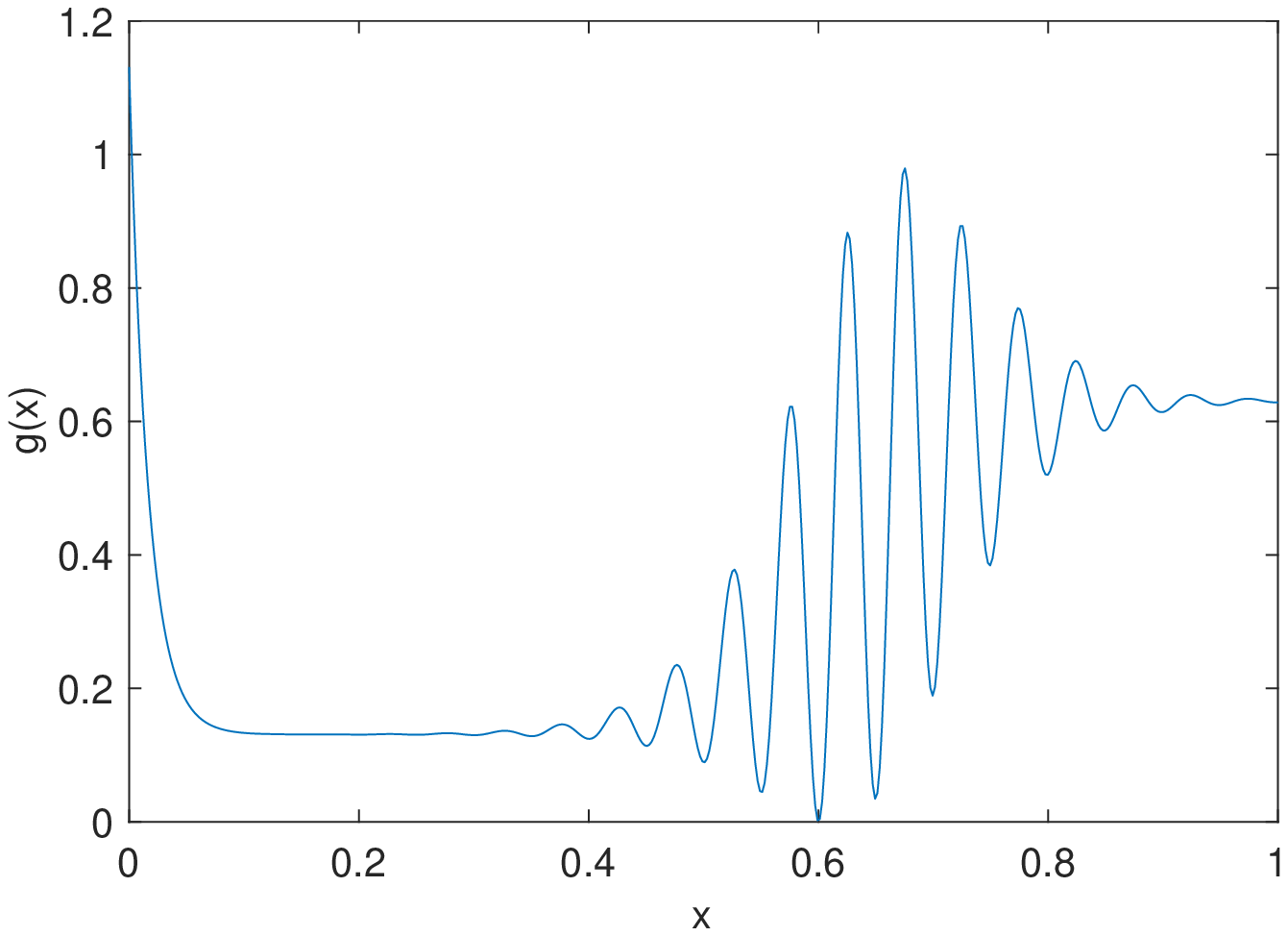}}
	\subfloat[\label{fig:g_robust} Effect of the presence of noise in a neighborhood of the global optimum and local stable optimum  values for the multi-modal function $g(\mathbf{x})$. ]{\includegraphics[width=0.5\linewidth]{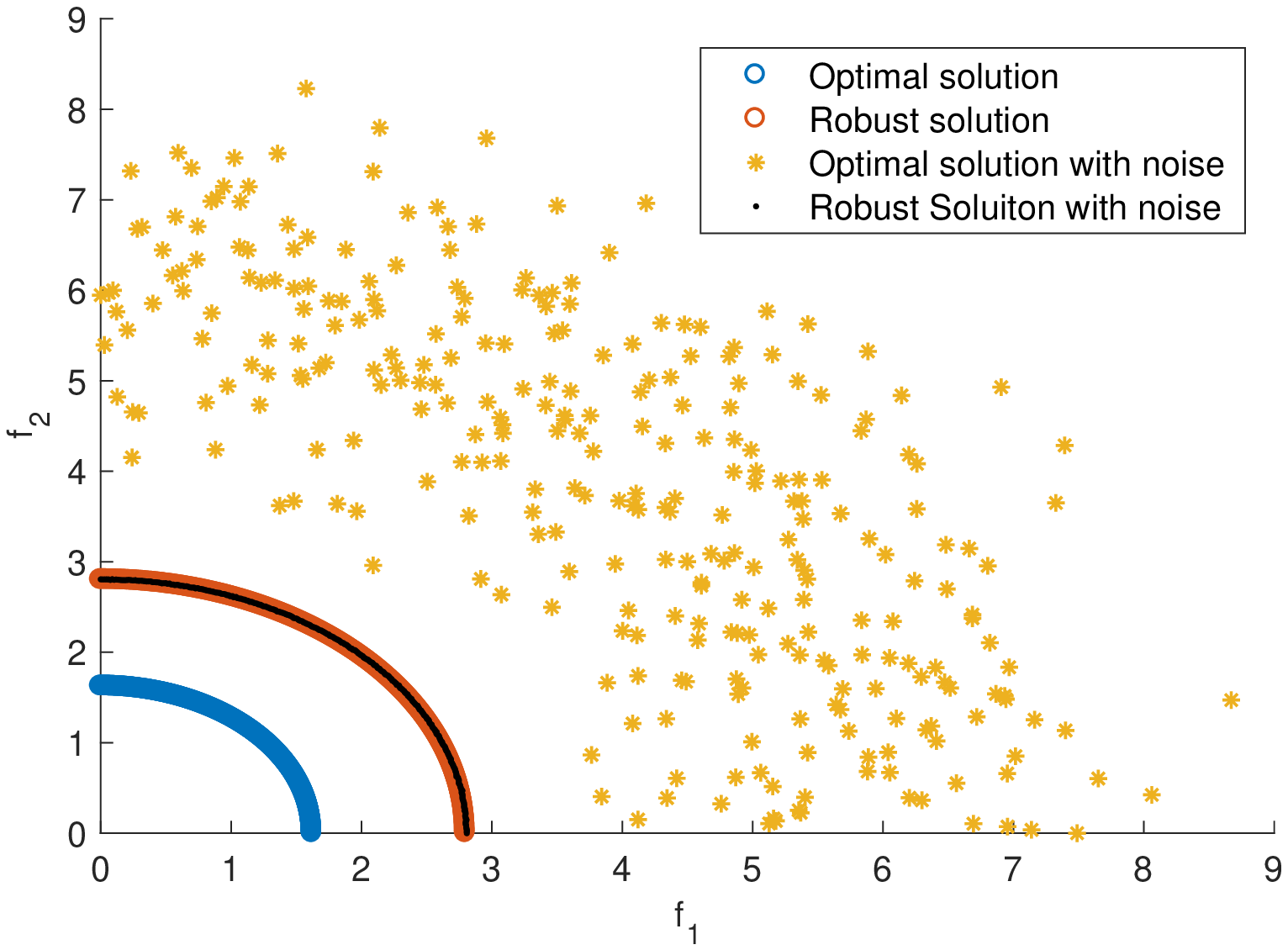}}
	\caption{Robust function $g(\mathbf{x})$}
\end{figure}

The function $g(\mathbf {x})$ is illustrated in Figure \ref{fig:fo_robust} for a single variable $x_1 \in [0, 1]$. Figure \ref{fig:g_robust} shows the effect caused by the presence of noise in $\mathbf{x}_d$ variables of the $g (\mathbf{x})$ function. The figure shows solutions obtained with the optimal value ($ x_i = 0.6 ~ \in \mathbf{x}_d $ -- blue dots) and the stable local optima  value ($ x_i = 0.2 \in (0.1, 0.3)) $ -- red dots) in the DTLZ2 problem with two objectives. To verify the robustness of these solutions, an $\epsilon$ noise with uniform distribution ($\epsilon \sim {U} (- 0.1,0.1)$) was added to the global optimum and in the stable local optimum. It is possible to see that when these solutions were re-evaluated, there were no changes in the distribution and convergence of the stable solutions (black dots), whereas the optimal solutions present unsatisfactory performance in the presence of noise (yellow dots). In this sense the stable local optimal solution is robust, while the global optimal solution is not robust. Note that using this function it is possible to create test problems where the robustness of the solutions can be tested in all objectives. Recent work on robust optimization uses test functions where robustness can be controlled for only one objective or uses common test problems \cite{GasparCunha.etal:2013, Meneghini2016, he2018robust, Mirjalili2018}.

%%=========================================
\subsection{Equality and Inequality Constraints}
\label{subsec_eq_ineq_constraintsrestricoes}

The proposed optimization problem $F (\mathbf{x})$ easily allows the incorporation of equality and inequality constraints with the manipulation of the $\phi (\mathbf{x})$ function defined previously in Eq. \eqref{eq:phi} and one vector $\mathbf{d}$. Since this function allows the spatial location of points in the objective space, the main idea is to select solutions with special values of $\phi(\mathbf{x})$ for a particular vector $\mathbf{d}$. Since $0 \leq \phi(\mathbf{x}) \leq 1$, we just select some thresholds $A,B$, with $0 < A < B <1$ and define the following constraints:

\begin{align}
\phi(\mathbf{x}) \geq A \label{eq:rstr_des_1}\\
\phi(\mathbf{x}) \leq A \label{eq:rstr_des_2}\\
A \leq \phi(\mathbf{x}) \leq B \label{eq:rstr_des_3}
\end{align}

In addition to the constraints presented in Eqs. \eqref{eq:rstr_des_1}, \eqref{eq:rstr_des_2} and \eqref{eq:rstr_des_3}, it is possible to select large regions in the objective space in the following way: consider a problem where the objective space is located in the first orthant of $\mathbb{R}^M$ space. In this case, consider the angle $\theta_i$ between the point $y = F (\mathbf{x})$ and the vectors of the canonical basis $ e_i = (0, \ldots ,1, \ldots, 0) $, with $ 1 \leq i \leq M $. Let $\theta_j = \min \{ \theta_1, \ldots, \theta_M \} $ and $ j $ the objective associated with this minimal value. The function $ \xi (\mathbf{x})$ associates the vector $\mathbf{x}$ to the objective $j$ which has the smallest angular distance of $F(\mathbf{x})$. It is easy to see that $\xi (\mathbf{x}) = j$ is not an injection function, because for $F (\mathbf{x}) = (1, \ldots, 1)$ we have $ \theta_1 = \theta_2 = \ldots \theta_M $, for example. In these cases where $ \min \{\theta_1, \ldots, \theta_M \} = \{\theta_{j1}, \ldots, \theta_{jk} \} $, make $\xi(\mathbf{x})= \min \{j1, \ldots, jk\}$ and define the constraint represented in Eq. \eqref{eq:rstr_igual_2}.

\begin{equation}\label{eq:rstr_igual_2}
\xi(\mathbf{x})=j
\end{equation}

\begin{figure}[!htb]
	\centering
	\subfloat[\label{fig:constrain_01} $\phi(\mathbf{x}) \geq 0.5$ and reference vectors $\mathbf{e}_1, \mathbf{e}_2$ and $\mathbf{e}_3$. ]{\includegraphics[width=0.4\linewidth]{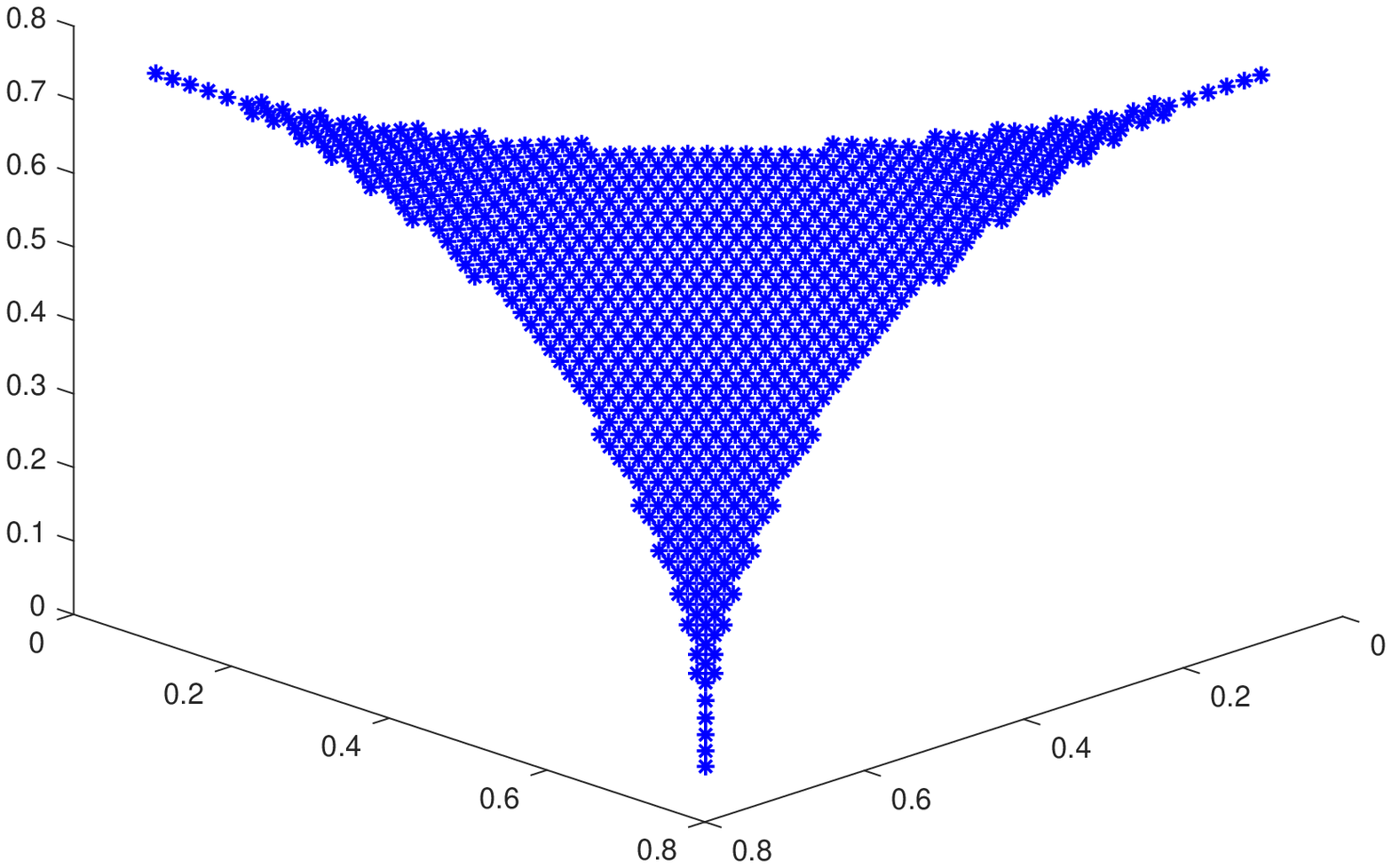}}
	\subfloat[\label{fig:constrain_02} $\phi(\mathbf{x}) \geq 0.5$ and reference vectors $\mathbf{e}_1, \ldots ,\mathbf{e}_5$ .]{\includegraphics[width=0.4\linewidth]{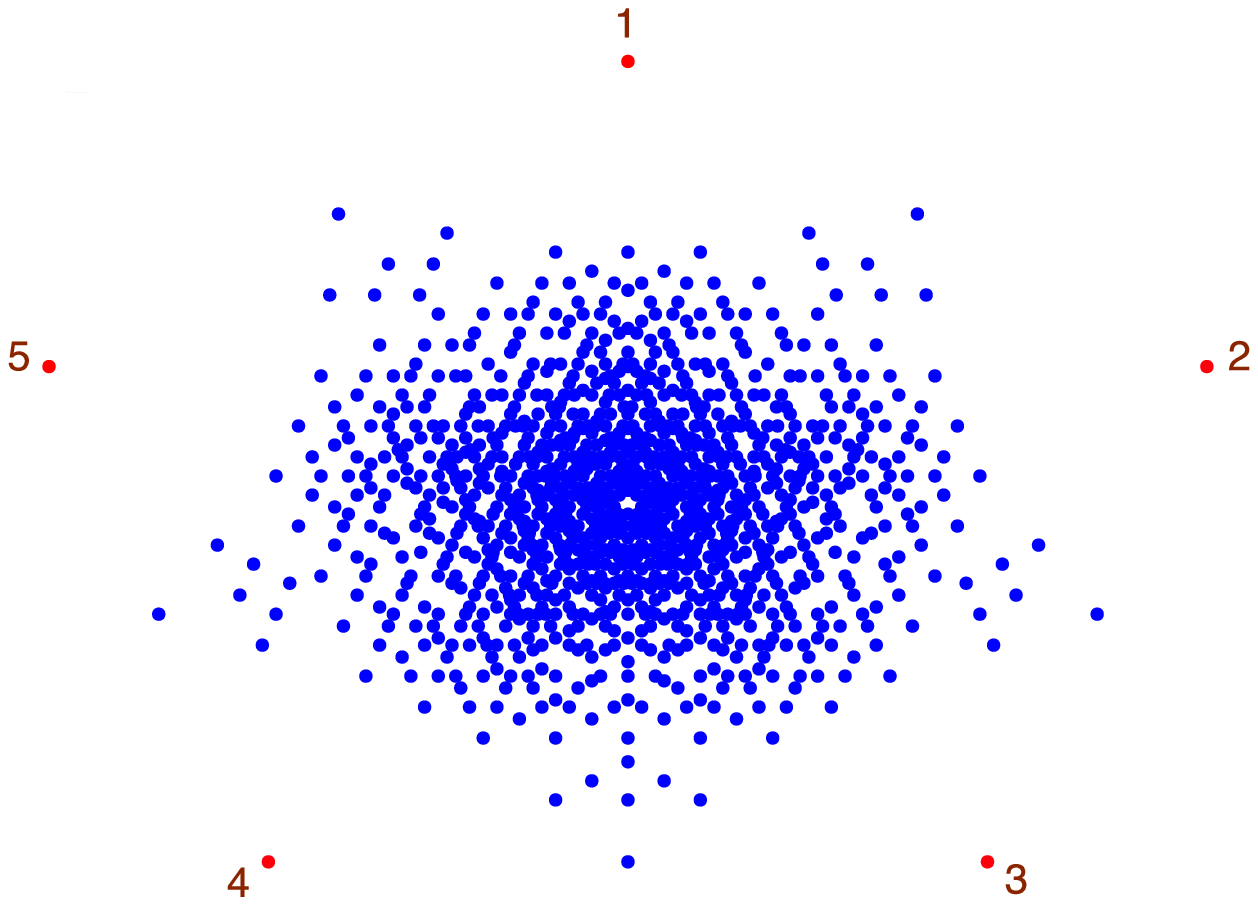}}\\
	\subfloat[\label{fig:constrain_03} $\phi(\mathbf{x}) \geq \{0.5,0.3,0.1\}$ and reference vectors $\mathbf{e}_1, \mathbf{e}_2$ and $\mathbf{e}_3$. ]{\includegraphics[width=0.4\linewidth]{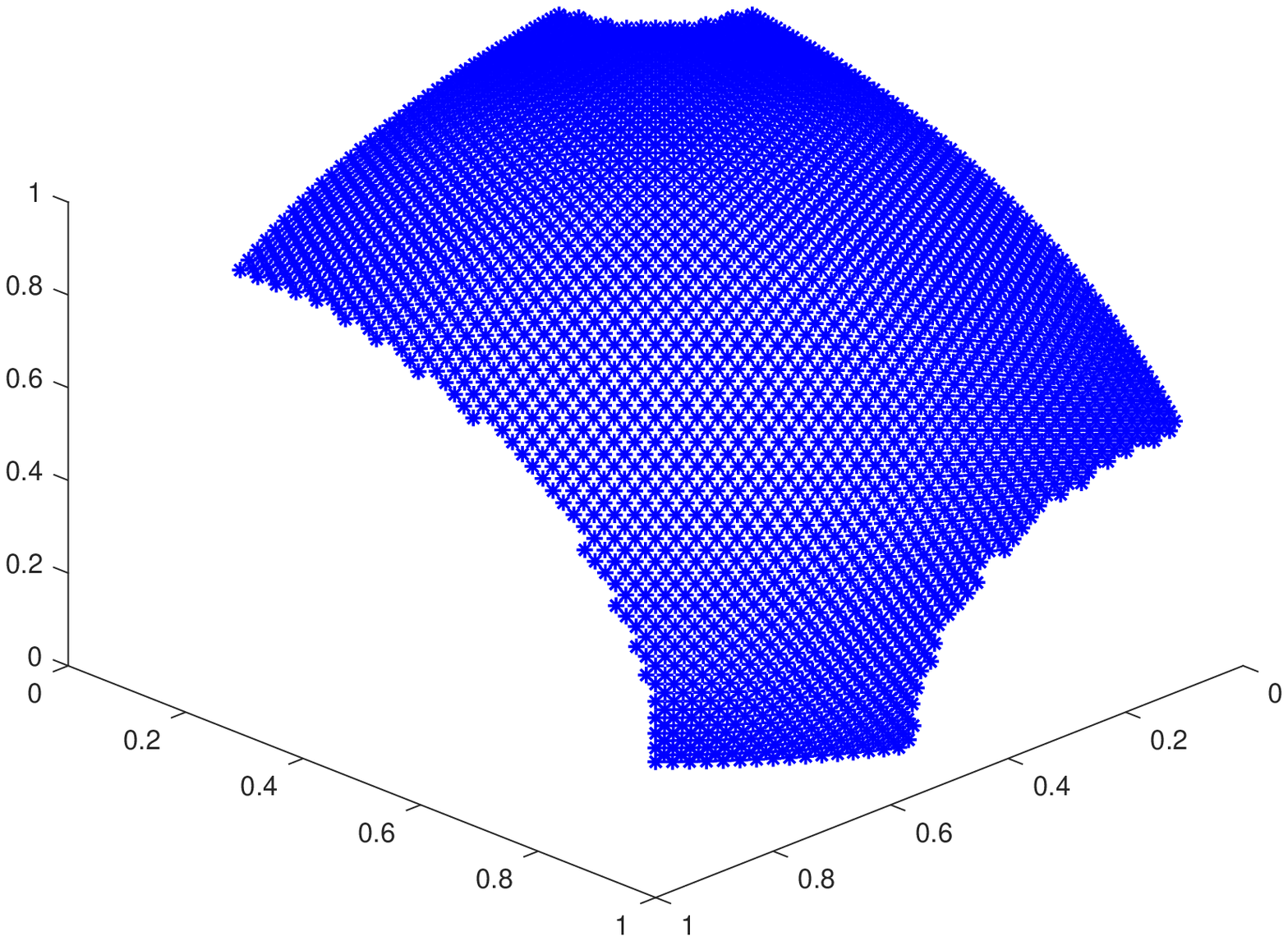}}
	\subfloat[\label{fig:constrain_04} $0.3 \leq \phi(\mathbf{x}) \leq 0.7$ and reference vector $\mathbf{d}=(1,1,1)$.] {\includegraphics[width=0.4\linewidth]{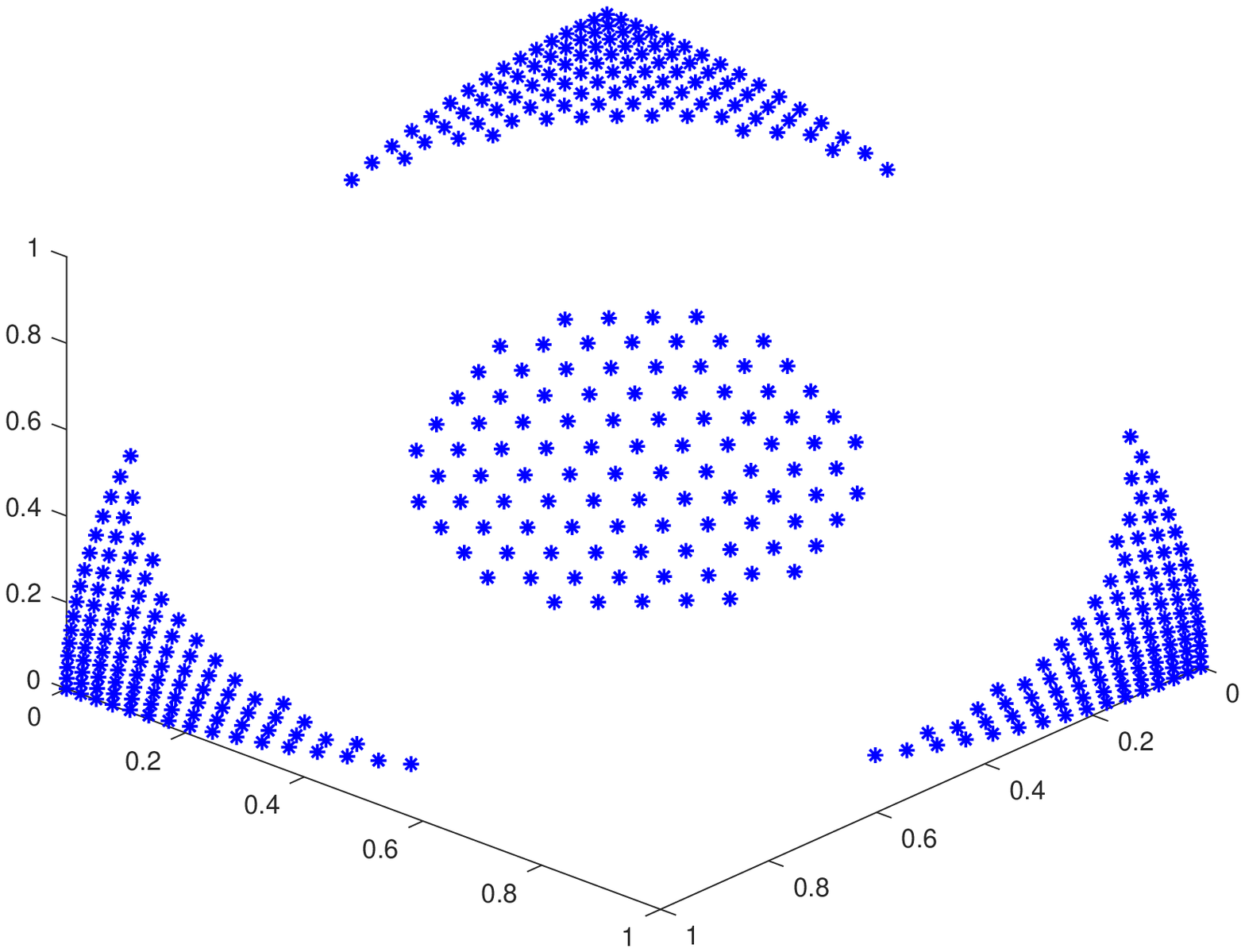}}\\
	\subfloat[\label{fig:constrain_05} $\xi(\mathbf{x})=2$]
	{\includegraphics[width=0.6\linewidth]{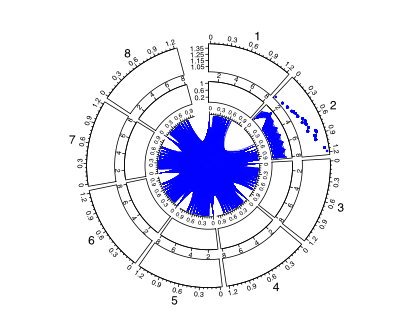}}
	\caption{Constrained Pareto Front using different reference vectors $\mathbf{d}$ and thresholds $A$ and $B$.} \label{fig:constrains}
\end{figure}

In this way, using this classification it is possible to select one or more regions that should be included or excluded from the Pareto front. All possibilities listed here can be incorporated into the MOP explicitly as constraints or as penalties in the distance function $F_d(\mathbf{x})$. Figure \ref{fig:constrains} ilustrates some examples. Figures \ref{fig:constrain_01} and \ref{fig:constrain_02} present the Pareto front of a three objective problem by applying the constraint defined in Eq. \eqref{eq:rstr_des_1}. Instead of a single reference vector $\mathbf{d}$, each canonical base vectors $\mathbf{e}_1=(1,0,0), \mathbf{e}_2=(0,1,0)$ and $\mathbf{e}_3=(0,0,1)$ were used as reference. Figure \ref{fig:constrain_01} uses $A=0.5$ while Figure \ref{fig:constrain_03} uses $A$ parameter equal to 0.5, 0.3 and 0.1 for the vectors $e_1, e_2$ and $e_3$ respectively. Notice that the resulting Pareto front is similar to an inverted Pareto front, but it is generated as a result of adding constraints. The same procedure is used in a problem with five objectives, using the canonical base vectors $\mathbf{e}_1, \ldots, \mathbf{e}_5$ and $A=0.5$. The Pareto front is shown in Figure \ref{fig:constrain_02} using the RadViz visualization tool \cite{radviz}. In this figure, the red points represent the vectors $\mathbf{e}_1, \ldots, \mathbf{e}_5$ just as reference. These examples illustrate how to use the constraint set to obtain a rich variety of shapes for the Pareto front. Figure \ref{fig:constrain_04} illustrates the use of the constraint defined by Eq. \eqref{eq:rstr_des_3} using $\mathbf{d}=(1,1,1), ~ A=0.3$ and $B=0.7$. This constraint can produce a disconnected Pareto front. Lastly, Figure \ref{fig:constrain_05} shows the constraint defined by Eq. \eqref{eq:rstr_igual_2} in a problem with eight objectives using the CAP-vis tool \cite{Meneghini2018,koochaksaraei2017new}. In this example, the Pareto front consists only of points near the axis of the second objective in terms of angular distance. For more details about reading this chart, see \cite{Meneghini2018}. For this case the constraint in Eq.  \eqref{eq:rstr_igual_2} was defined as $\xi(\mathbf{x})=2$.

%%=========================================
\section{Proposed Generator of benchmark problems}
\label{sec_proposedfunctions}

This section presents a new test function generator for scalable and customizable benchmark problems in MaOPs. This new set uses the bottom-up approach and allows the creation of scalable problems for any number of objectives, presenting Pareto fronts with different shapes and topologies.

The auxiliary functions previously defined allow the formulation of several customizable optimization problems. We refer to it as the Generalized Position-Distance (GPD) test functions. The generated functions are all scalable and use either the deceptive version, Eq.~\eqref{eq:g_deceptiva}, or the multimodal version, Eq.~\eqref{eq:g_robusta}, of function $g(\mathbf{x})$. The dissimilarity of the functions can also be controlled.  In addition, each GPD test problem can present any of the constraints described in Section \ref{subsec_eq_ineq_constraintsrestricoes}.

All the functions generated use the bottom-up approach in the multiplicative or additive form, as described in Eq. \eqref{eq:POM}.

\begin{align} \label{eq:POM}
\min F(\mathbf{x}), ~& F: \mathbb{R}^N  \to \mathbb{R}^M \nonumber\\  
\mbox{subject to } &
\begin{cases}
\phi_i(\mathbf{x})\leq A_i, & i=1 \ldots k_1 \\
\xi_j(\mathbf{x}) = I_j, & j=1
\ldots k_2, ~ I_j\in \{1 \ldots M\}
\end{cases}
\end{align}

The decision space $\mathbb{R}^N = \mathbb{R}^R \times \mathbb{R}^S$ is separable, being the vectors $\mathbb{R}^R$ responsible for the relative position of $F(\mathbf{x})$ in the objective space (position parameters) and $\mathbb{R}^S$ for the convergence of points in the Pareto front (distance parameters). The space $\mathbb{R}^R$ has $r = (M-1) q + t$ coordinates, where $M$ is the number of objectives and $q, t$ are the parameters of the meta-variable $y$ defined by Eq. \eqref{eq:meta_variavel}, with $2t+1<q$. If $q=1$ and $t=0$ then $y_i=x_i$ and Eq. \eqref{eq:T} is the usual multidimensional polar coordinates. Then, to define any problem instance, it is necessary to specify the number of objectives $M$. The number of variables of the decision space is given by $N=(M-1)q + t + S$, where $q$ and $t$ are the meta variables parameters and $S$ is the number of variables used in some distance functions. If the meta-variable is used then $\mathbf{x}_p \in [-1,1]^R$ and $\mathbf{x}_d \in [0,1]^S$ for the distance functions presented in this paper.

$F_p(\mathbf{x})$ defines the relative position of points in the objective space and the p-norm (or quasi-norm) of the Pareto front. This function uses the parameters $q, t$ (previously defined) and $p$, the latter being used to normalize the Pareto front, affecting its shape.

Points in a constant norm surface in high dimensional space have unbalanced coordinates. For example, in space with $M$ dimensions, a vector $\mathbf{e}_i = (0, \ldots, 0,1,0, \ldots, 0)$ with canonical base has $p$-norm equal to $1$ for any value of $p>0 $. Points located on the edge of the first orthant of the space $\mathbb{R}^M$ have constant $p$-norm equal to $1$. On the other hand, in an extreme case, a vector $ \mathbf{v} $ parallel to the hyper-diagonal $ \mathbf {d} = (1, \ldots, 1) $ with $p$-norm equal to $1$ has the following coordinates:

\begin{align}
\mathbf{v}& = \frac{\mathbf{d}}{||\mathbf{d}||_p} \\ 
& = \left(\dfrac{1}{\sqrt[p]{M}}, \ldots, \dfrac{1}{\sqrt[p]{M}} \right) 
\end{align}

\begin{figure}[!htb]
	\centering
	\includegraphics[width=0.85\linewidth]{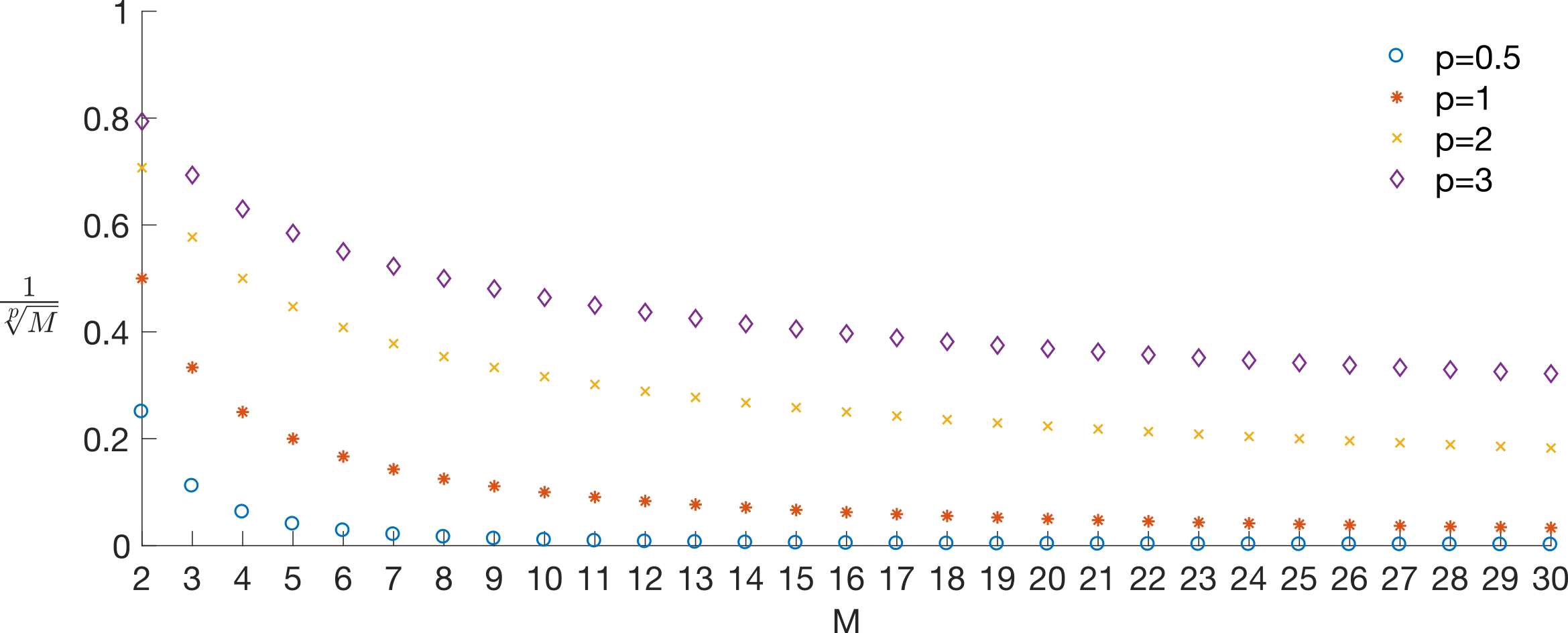}
	\caption{$\frac{1}{\sqrt[p]{M}}$ on $M$-dimensional spaces}
	\label{fig:coordenadas_pnorma}
\end{figure}

For a fixed $p$ value, $ \frac{1}{\sqrt [p] {M}} $ decreases very quickly. Figure \ref{fig:coordenadas_pnorma} exemplifies the evolution of the values of $\frac{1}{\sqrt [p] {M}} $ related to the space dimension $ M $. It is possible to realize that this value decreases with $ M $ reaching values close to zero in high dimensionality. It can be solved easily by determining an ideal value for $p $. As an example, use $ p> \frac{\ln(M)}{\ln (2)}$ to obtain $ \frac{1}{\sqrt[p]{M}}>\frac{1}{2} $. In this way, for MaOPs, we suggest $ p = \left\lceil \frac{\ln(M)}{\ln(2)} \right \rceil $.

$F_d(\mathbf{x})$ establishes the radial distance between the points in the objective space and the Pareto front using some auxiliary function. The literature presents many examples of such functions, with distinct characteristics \cite{wang2018generator,wang2018scalable,Deb2005DTLZ}. The auxiliary functions introduced in this paper present the following peculiarities:
\begin{enumerate}
    \item Use of the relative position of a point in the objective space by the $\phi(\mathbf{x})$ function correlating position and distance. With this feature, changing the relative position by changing the $\mathbf{x}_p$ vector changes the distance function value to the same $\mathbf{x}_d$ vector; 
    \item Deceptiveness: a large portion of the decision space leads to a suboptimal distance value while optimal distance values are restricted to a small region of this space;
    \item Robustness: An optimal distance solution is sensitive to small disturbances while suboptimal solutions have more stability in the presence of noise in distance variables. The proposed  auxiliary function makes a Robust Multi-objective Optimization Problem where the robustness can be analyzed on all objectives. In a recent paper, \citet{he2018robust} present a Robust Multi-objective Evolutionary Algorithm but the test function used enables analysis of robustness of solutions in just one objective.
\end{enumerate}

The deceptive auxiliary function uses a parameter $k$ in Eq. \eqref{eq:r} that defines the number of large and narrow valleys. The robust auxiliary function has no parameters. Since the smallest value of each of its components in both functions is equal to zero, the corresponding function $F_d(\mathbf{x})$ is $g(\mathbf{x})$ in the additive approach and $1 + g(\mathbf{x})$ in the multiplicative case.

\begin{figure}
	\centering
	\subfloat[\label{fig:xtra_FD01_2obj} Distance function \eqref{eq:mixedFD} with 2 objectives. ]{\includegraphics[width=.5\linewidth]{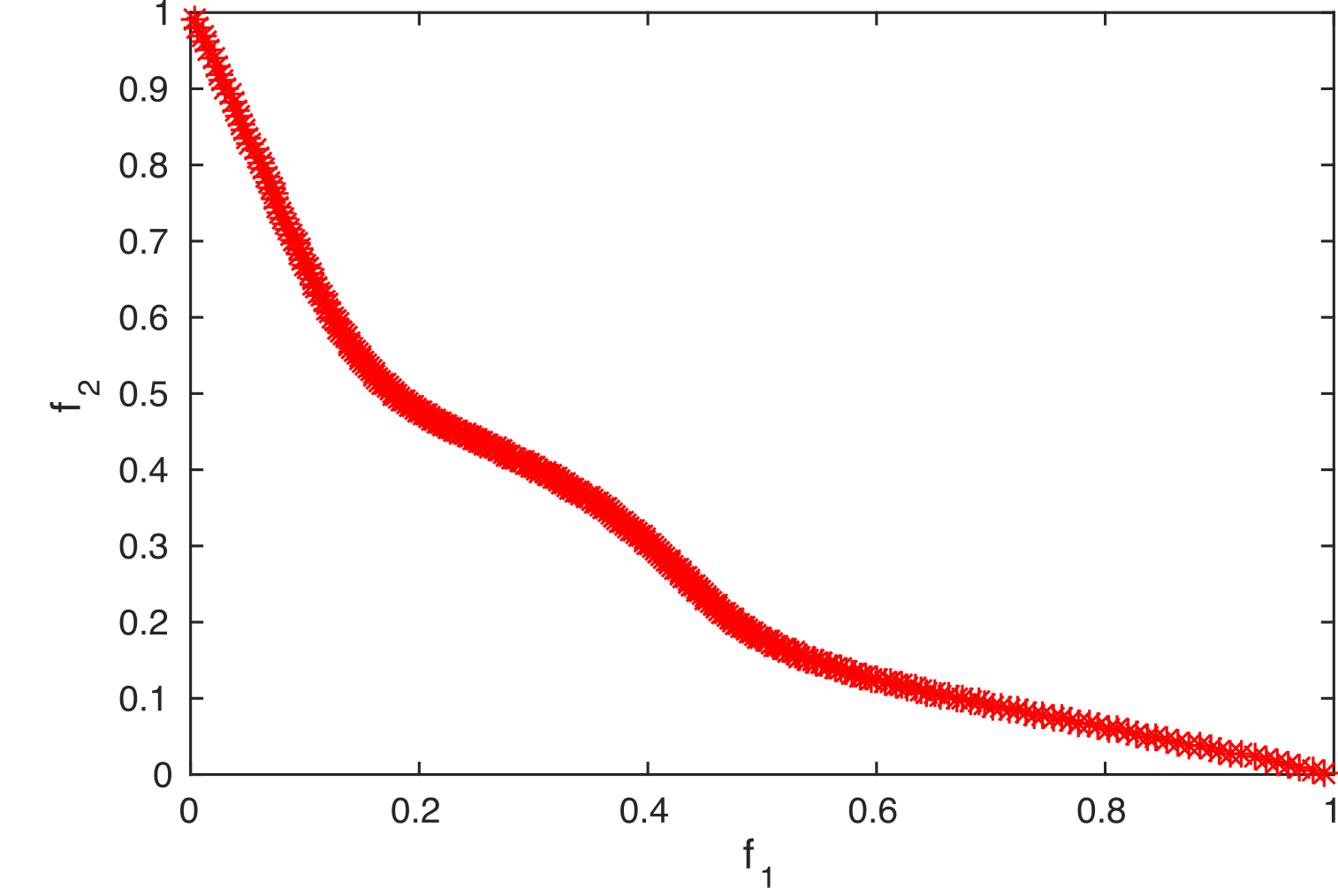}}
	\subfloat[\label{fig:xtra_FD01_3obj} Distance function \eqref{eq:mixedFD} with 3 objectives. ]{\includegraphics[width=.5\linewidth]{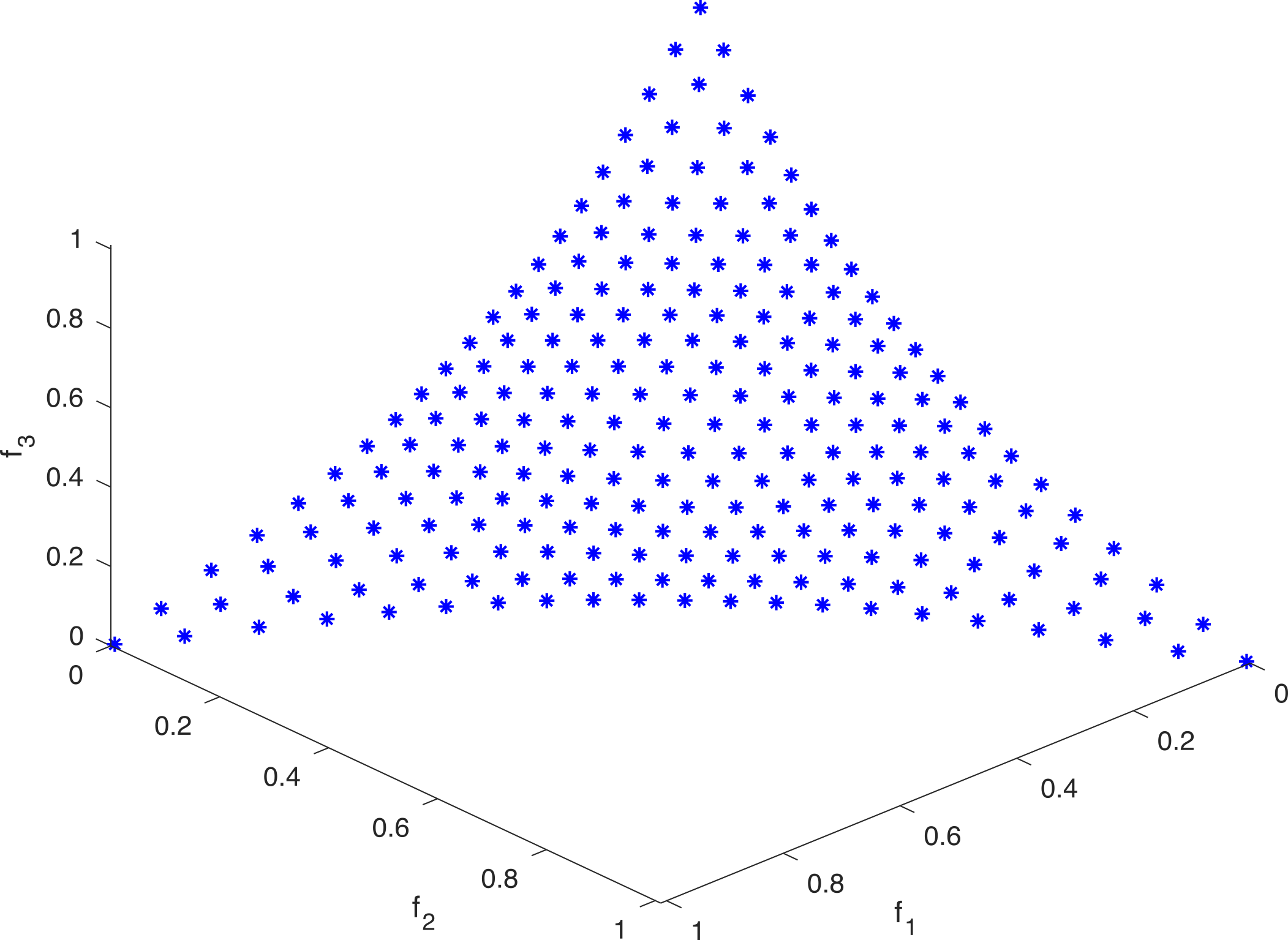}} \\
	\subfloat[\label{fig:xtra_FD02_2obj} Distance function \eqref{eq:xtraFD02} with 2 objectives. ]{\includegraphics[width=.5\linewidth]{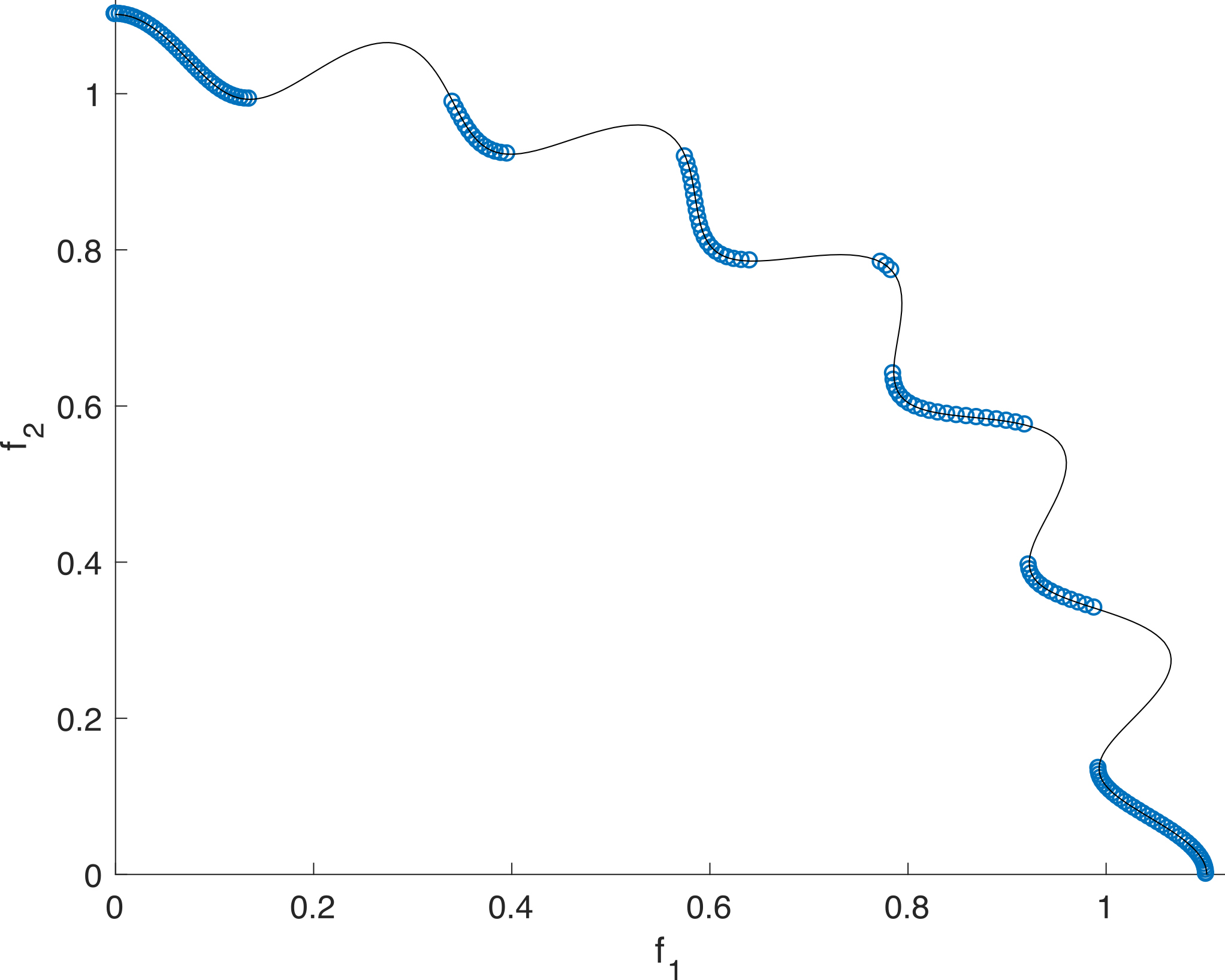}}
	\subfloat[\label{fig:xtra_FD02_3obj} Distance function \eqref{eq:xtraFD02} with 3 objectives. ]{\includegraphics[width=.5\linewidth]{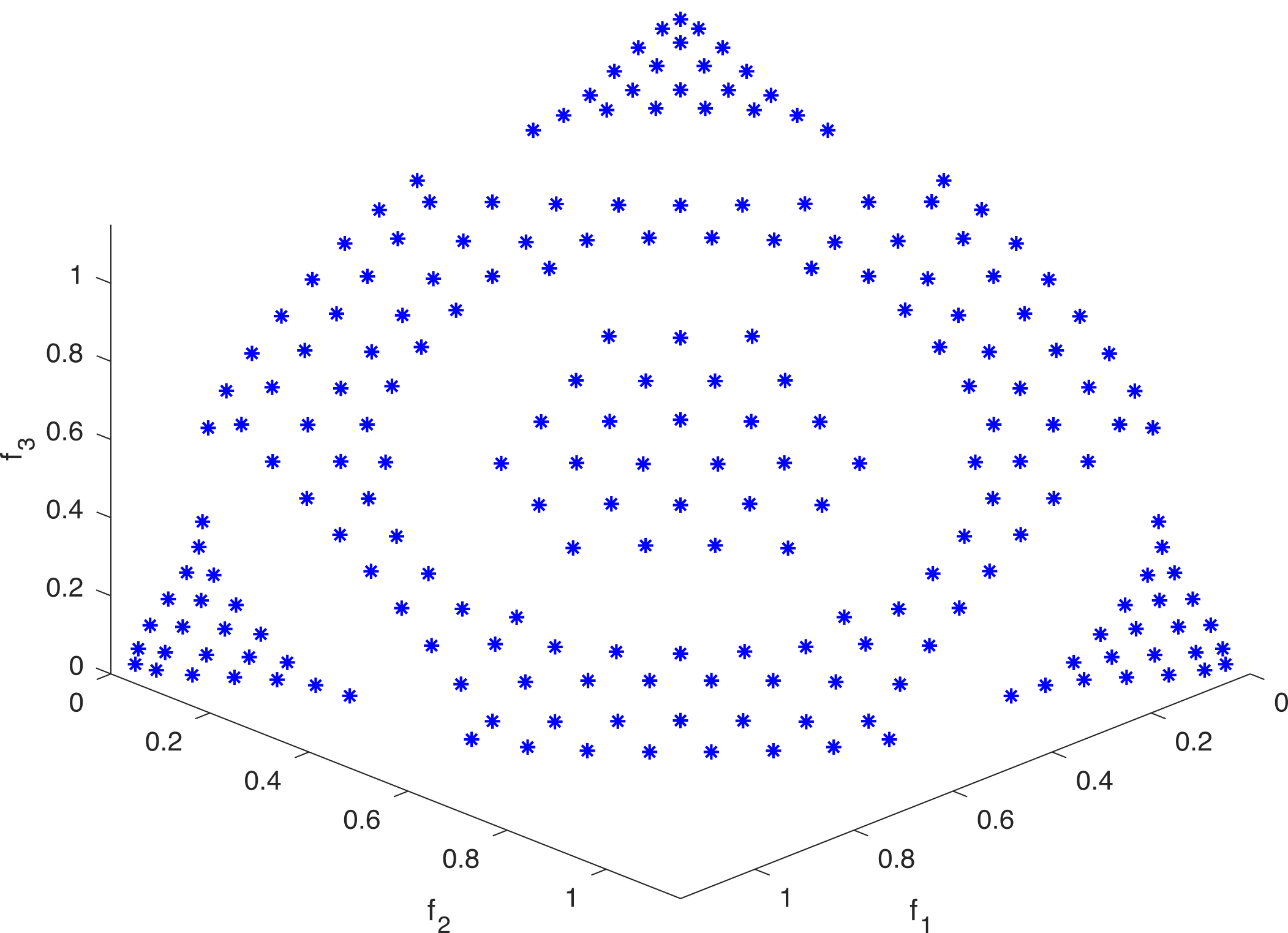}}
	\caption{Pareto front of the distance function \eqref{eq:mixedFD} and \eqref{eq:xtraFD02} with 2 e 3 objectives}
	\label{fig:xtra_FD01_FD02}
\end{figure}

In addition to the $F_d(\mathbf{x})$ functions listed above, it is possible to create other functions by manipulating the proposed $g(\mathbf{x})$ distance and $\phi(\mathbf{x})$ functions. For example,

\begin{equation} 
	F_d(\mathbf{x}) = \frac{\phi(\mathbf{x})^5}{2} + g(\mathbf{x}) + 0.5
	\label{eq:mixedFD}
\end{equation} 
presents a Pareto front with convex and concave regions, which are symmetrical to the hyper-diagonal $\mathbf{d} = (1,1, \ldots, 1)$ of the first orthant of the objective space. Figure \ref{fig:xtra_FD01_2obj} and \ref{fig:xtra_FD01_3obj} show the Pareto front of this problem with two and three objectives.

A function with disconnected Pareto front and symmetric with the hyper-diagonal $\mathbf{d} = (1, \ldots, 1)$ of the first orthant in the objective space can be produced by using:
\begin{equation}
	F_d(\mathbf{x}) = \frac{\cos(3\pi \phi(\mathbf{x}))^2}{10} + g(\mathbf{x}) + 1
    \label{eq:xtraFD02}
\end{equation}
Figure \ref{fig:xtra_FD02_2obj} and \ref{fig:xtra_FD02_3obj} show the Pareto front of this problem with two and three objectives.

In addition to the possibilities listed above, the constraints presented in Eqs. \eqref{eq:rstr_des_1} to \eqref{eq:rstr_igual_2} can be added, as well as the dissimilarity of objectives as the presented in Eq. \eqref{eq:D}.

This proposal for generating test problem instances focuses on the combination of the parameters presented for the composition of the meta-variable $y$, the position function $F_p(\mathbf{x})$ and the distance function $F_d(\mathbf{x})$, as well as the combined use of one or more constraints. The composition of these parameters is capable of producing an unlimited number of multi-objective, multi-purpose, large-scale optimization benchmark problems that can be multimodal, deceptive, dissimilar, constrained, and others.

\begin{table}[htb]
\centering
\caption{List of parameters for generative benchmarking}
\label{tab:generative_benchmarking}
\begin{tabular}{llp{3.5cm}p{7.5cm}}
	\toprule
	Parameter & Equation & Feature & Notes \\
	\midrule
	$M$ & & Number of objectives   & $M\geq 2$ \\ \midrule
	$N$ & & Number of decision variables & $N= R + S $ where $F_p(\mathbf{x}):\mathbb{R}^R \to \mathbb{R}^M$ and $F_d(\mathbf{x}):\mathbb{R}^S \to \mathbb{R}$ \\ \midrule
	$q,t$ & \eqref{eq:meta_variavel} & Meta-variable $y_i$, bias, separability. & Defines the length and the overlap size of the meta-variable $y_i$ and the dimension $R=(M-1)q+t$ of $\mathbb{R}^R$ space. Setting $q=1$ and $t=0$ get the usual $M-1$ position variables. $2t+1 < q$.\\  \midrule
	$p$	& \eqref{eq:p-norm}	& $p$-norm value, shape of the Pareto front.& Defines the $p$-norm value used in the $F_p(\mathbf{x})$ function, controlling the convexity or concavity of the front. $p>0$ \\ \midrule
	%$\phi(\mathbf{x})$ & \eqref{eq:phi}	& Normalized angular distance function & Used in some $F_d(\mathbf{x})$ distance function and in the constrains. \\ \midrule
	$\mathbf{d}$ &	& Reference vector & Used in $\phi(\mathbf{x})$ function, see \eqref{eq:phi}. \\ \midrule
	$k$ & \eqref{eq:r}& Valley width & Defines the number of narrow and wide valleys in the $r(\mathbf{x})$ function.\\ \midrule
	$D(\mathbf{x})$ & \eqref{eq:D} & Dissimilar Objectives & Performs the PF transformation from $0 \leq f_i(\mathbf{x}) \leq 1$ to $-2i \leq f_i(\mathbf{x}) \leq 2i$ \\ \midrule
	$g(\mathbf{x})$ & & Auxiliary function & Defines the problem as being deceptive  \eqref{eq:g_deceptiva}, multimodal, or having robust solutions \eqref{eq:g_robusta},  \\ \midrule
	$+, ~ \times$ & \eqref{eq:POM} &  & Defines the problem as additive or multiplicative \\ \midrule
	$A, ~ B$ & \eqref{eq:rstr_des_1} to \eqref{eq:rstr_des_3} & Inequality constrains & Defines the inequality constrains. $0 \leq A < B \leq 1 $ \\ \midrule
	$j$ & \eqref{eq:rstr_igual_2}  & Equality constrains & Defines the equality constrains $\xi(\mathbf{j})$. $0 \leq j \leq M$\\ \midrule
	\bottomrule
\end{tabular}
\end{table}

Table \ref{tab:generative_benchmarking} presents the main parameters of the proposed instance generator. Suppose as an example that a  robust problem with the following characteristics is required:
\begin{enumerate}
    \item Two objectives and several decision variables;
    \item Convex, dissimilar objectives and disconnected Pareto front;
    \item Multiplicative approach.
\end{enumerate}

This specific problem can be produced by setting the following parameters:
\begin{enumerate}
    \item Set $M=2$ and use the meta-variables in Eq.  \eqref{eq:meta_variavel} with $q=10$, $t=4$ and $S=15$, generating the minimization problem $F:\mathbf{R}^{29} \to \mathbf{R}^2$.
    \item Set $p=2$ and use $D(\mathbf{x})$ in $F_p(\mathbf{x})$ function. Include the $ \phi(\mathbf{x}) \geq 0.3$ and $\phi(\mathbf{x}) \leq 0.7$ constraints, using $\mathbf{d}=(1,1)$. Other values for $A$ can be used instead 0.3 and 0.7, as well as for the vector $\mathbf{d}$ and any other value greater than 1 for parameter $p$.
    \item Define $F(\mathbf{x})=F_p(\mathbf{x})F_d(\mathbf{x})$ where $F_d(\mathbf{x})=1+ g(\mathbf{x})$ and $g(\mathbf{x})$ is defined by Eq. \eqref{eq:g_robusta}.
\end{enumerate}

A deceptive MaOP for which the nominal values of the objectives are not discrepant but the first objective is always higher than the others can be created by making $M > 3,~ p=\bigl\lceil \frac{\log(M)}{\log(2)}\bigr\rceil$, using the $g(\mathbf{x})$ deceptive auxiliary function and the $\xi(\mathbf{x})=1$ constraint.

The following is a quick roadmap for building test function instances.

\begin{description}
\item[Input:] Number of Objectives $M$, number of distance variables $S$, reference vector $\mathbf{d}$, a list of parameters $q$, $t$, $p$, $k$, and other features that should be imposed on the problem (deceptiveness, dissimilarity, separability, constraints etc);
\item[Decision Space design:] Initialize the Decision Space. If meta-variable $y_i$ is used, set $R=(M-1)q+t$, else do $R=M-1$. Then generate the decision vector $\mathbf{x}=(\mathbf{x}_p,\mathbf{x}_d) \in \mathbb{R}^{R+S}$ where $\mathbf{x}_p \in [-1,1]^R$ and $\mathbf{x}_d \in [0,1]^S$.
\item[Evaluates the position function $F_p(\mathbf{x})$:] If the meta-variable is used, evaluate Eq. \eqref{eq:meta_variavel}
$$
y_i= \frac{1}{q+t}\left|\sum_{j=(i-1)q+1}^{iq+t} x_j\right| , -1 \leq x_j \leq 1
$$
Else do $y_i=|x_i|, ~ i=1 \ldots M-1$. Then project $\mathbf{x}_p$ into space $\mathbb{R}^M$ using the spherical coordinates in Eq. \eqref{eq:T}
\begin{equation*}
T(\mathbf{x})=
\left\lbrace 
\begin{aligned}
t_1(\mathbf{x})	   & =\cos(y_1 \pi/2)\cos(y_2 \pi/2) \ldots \cos(y_{M-2}\pi/2)\cos(y_{M-1}\pi/2)\\
t_2(\mathbf{x})	   & =\cos(y_1 \pi/2)\cos(y_2 \pi/2) \ldots \cos(y_{M-2}\pi/2)\sin(y_{M-1}\pi/2)\\
t_3(\mathbf{x})	   & =\cos(y_1 \pi/2)\cos(y_2 \pi/2) \ldots \sin(y_{M-2}\pi/2)\\
\vdots			   & \\
t_{M-1}(\mathbf{x})& =\cos(y_1 \pi/2)\sin(y_2 \pi/2)\\
t_M(\mathbf{x})	   & =  \sin(y_1 \pi/2)	
\end{aligned} 
\right.
\end{equation*}
Evaluate the $p-$norm $||T(\mathbf{x})||_p$ in Eq. \eqref{eq:p-norm}
\begin{equation*}
||T(\mathbf{x})||_p = h(\mathbf{x})=\left( \sum_{i=1}^{M} |t_i(\mathbf{x})|^p \right)^{1/p}
\end{equation*}
and normalize $T(\mathbf{x})$ by defining the Eq. \eqref{eq:F_d}
$$F_p(\mathbf{x})=\frac{T(\mathbf{x})}{h(\mathbf{x})}$$
Additionally, if the problem is dissimilar, calculate Eq. \eqref{eq:D}
\begin{equation*}
D(\mathbf{x})=
\left\lbrace 
\begin{aligned}
d_1(\mathbf{x})	   & =2(2f_1(\mathbf{x}) - 1)\\
d_2(\mathbf{x})	   & =4(2f_2(\mathbf{x}) - 1)\\
d_3(\mathbf{x})	   & =6(2f_3(\mathbf{x}) - 1)\\
\vdots			   & \\
d_M(\mathbf{x})	   & = (2M)(2f_M(\mathbf{x}) - 1)
\end{aligned} 
\right.
\end{equation*}
and do $F_p(\mathbf{x})=D(\mathbf{x})$.
\item[Evaluate the distance function $F_d(\mathbf{x})$:] If function $\phi(\mathbf{x})$ is used in any step, evaluate the angle $\varphi(\mathbf{x})$  between vector $\mathbf{d}$ and $F_p(\mathbf{x})$ (Eq. \eqref{eq:angulo}) and the normalized angular distance $\phi(\mathbf{x})$ (Eq.\eqref{eq:phi}) using
\begin{align*} 
\varphi(\mathbf{x}) & = \arccos\left( \frac{\mathbf{d} \cdot F_p(\mathbf{x})^T }{|\mathbf{d}| |F_p(\mathbf{x})|} \right) \\
\phi(\mathbf{x}) & = \frac{\varphi(\mathbf{x})}{\varphi_{\max}}   
\end{align*}
Note that in the first orthant the maximum angle $\varphi_{\max}$ is the angle between the vector $\mathbf{d}$ and one canonical basis $\mathbf{e}_i$. In case of a dissimilar problem, this angle must be calculated before applying $D(\mathbf{x})$ function.
Select one auxiliary function $g(\mathbf{x})$ (Eqs. 	 \eqref{eq:g_deceptiva}, \eqref{eq:g_robusta}, \eqref{eq:mixedFD}, \eqref{eq:xtraFD02} or any other appropriate equation of your choice) and build the appropriate distance function $F_d(\mathbf{x})$. 
\item[Define the constrains:] Using one or more inequality or equality constraint, for e.g. Eqs. \eqref{eq:rstr_des_1}, \eqref{eq:rstr_des_2}, \eqref{eq:rstr_des_3} or \eqref{eq:rstr_igual_2}.
\end{description}

The proposed instance generator has great flexibility. Its modular structure allows for multipurpose problem creation. Its basic structure can be extended by adding new auxiliary functions $g(\mathbf{x})$ and $\phi(\mathbf{x})$, as well as new constraints or other meta-variable $y$.

\section{Conclusions}
\label{sec_conclusions}

Benchmark functions are an important validation tool for optimization algorithms and for driving research and development in computational intelligence. The functions analyzed in this paper showed many virtues. The proposed procedure for generating benchmark problems is able to maintain the qualities of the test problems presented before and fits the desirable characteristics exposed in the literature. The functions are easy to implement and interpret, present known Pareto solutions, and are scalable to any number of objectives.  A variety of Pareto shapes and topologies bring this proposed optimization problem closer to real problems and brings more challenges in the development of new algorithms. 

Another great advantage of the proposed GPD benchmark generator is the possibility of creating optimization problems for robust optimization with several characteristics, which is missing from existing test suites and it is becoming a major gap in the specialized literature. Along with this, it is possible to set equality and inequality constraints which have a clear and easy formulation and interpretation. By removing regions of the Pareto front a large variety of Pareto shapes and geometries can be produced.

It is known that obtaining optimal solutions in MaOPs is a very hard task. Although these functions are scalable to any number of objectives and present different shapes of Pareto fronts, a visualization task of the solutions may require some effort from the DM.
The proposed generative testing approach can be extended to support other features of additional classes of problems.

Finally, to publicize the generative benchmarking method proposed, it is interesting to include it in platEMO, a MATLAB platform for MOEA, as highlighted recently by \citet{tian2017platemo}, or as a Python library. 

\section{Acknowledgments}

This work has been supported by the Brazilian agencies (i) National Council for Scientific and Technological Development (CNPq); (ii) Coordination for the Improvement of Higher Education (CAPES) and (iii) Foundation for Research of the State of Minas Gerais (FAPEMIG, in Portuguese).

%%=========================================
%%=========================================
\section*{References}
\bibliographystyle{elsarticle-num-names}
\bibliography{referencias}

\end{document}